%% file: main.tex
\documentclass[10pt,twocolumn,letterpaper]{article}

\usepackage[pagenumbers]{iccv} 


\usepackage{lipsum}
\usepackage{caption}
\usepackage{subcaption}
\usepackage{xcolor}
\usepackage{soul}
\usepackage{tikz}
\usepackage[misc]{ifsym}

\def\eg{\emph{e.g.}} 
\def\ie{\emph{i.e.}}

\def\etal{\emph{et al.}}

\definecolor{generative}{RGB}{101,120,94}
\newcommand{\generative}[1]{\textbf{\textcolor{generative}{#1}}}

\definecolor{transformation}{RGB}{120,92,117}
\newcommand{\transformation}[1]{\textbf{\textcolor{transformation}{#1}}}

\definecolor{discriminative}{RGB}{81,110,122}
\newcommand{\discriminative}[1]{\textbf{\textcolor{discriminative}{#1}}}

\definecolor{iccvblue}{rgb}{0.21,0.49,0.74}
\usepackage[pagebackref,breaklinks,colorlinks,allcolors=iccvblue]{hyperref}

\def\paperID{3234} 
\def\confName{ICCV}
\def\confYear{2025}

\title{Is Visual in-Context Learning for \\ Compositional Medical Tasks within Reach?}

\author{Simon Rei{\ss}$^{1~\text{\Letter}}$ $\phantom{0000}$ Zdravko Marinov$^1$ $\phantom{0000}$ Alexander Jaus$^1$ $\phantom{0000}$ Constantin Seibold \\
M. Saquib Sarfraz$^{1, 2}$ $\phantom{0000}$ Erik Rodner$^3$ $\phantom{0000}$ Rainer Stiefelhagen$^1$ \\
{\small $^1$Karlsruhe Institute of Technology~ $^2$Mercedes-Benz Tech Innovation~ $^3$University of Applied Sciences Berlin} \\
{\tt\small \Letter~{simon.reiss@kit.edu}}
}

\begin{document}
\maketitle

\begin{abstract}
In this paper, we explore the potential of visual in-context learning to enable a single model to handle multiple tasks and adapt to new tasks during test time without re-training. Unlike previous approaches, our focus is on training in-context learners to adapt to sequences of tasks, rather than individual tasks. 
Our goal is to solve complex tasks that involve multiple intermediate steps using a single model, allowing users to define entire vision pipelines flexibly at test time. To achieve this, we first examine the properties and limitations of visual in-context learning architectures, with a particular focus on the role of codebooks.
We then introduce a novel method for training in-context learners using a synthetic compositional task generation engine.
This engine bootstraps task sequences from arbitrary segmentation datasets, enabling the training of visual in-context learners for compositional tasks.
Additionally, we investigate different masking-based training objectives to gather insights into how to train models better for solving complex, compositional tasks.
Our exploration not only provides important insights especially for multi-modal medical task sequences but also highlights challenges that need to be addressed.
\end{abstract}

\section{Introduction}
\label{sec:intro}
\begin{figure}
    \centering
    \includegraphics[width=\linewidth,page=8]{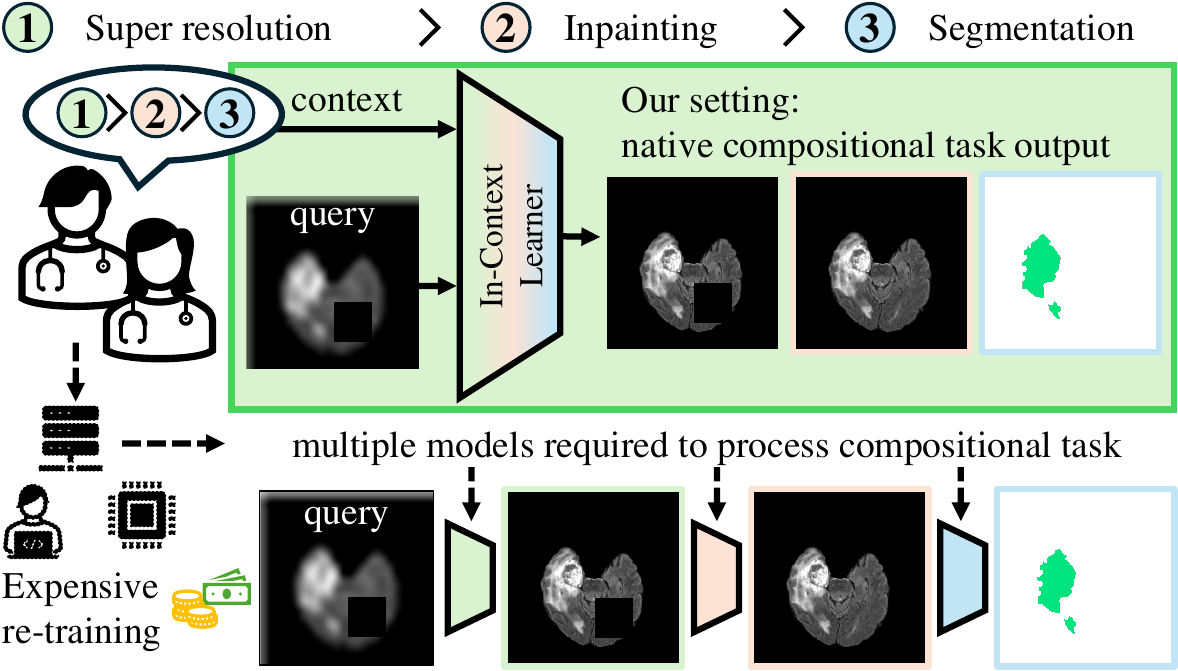}
    \caption{\underline{Top}: In green, our scenario of visual in-context learning for compositional tasks using one general model. \underline{Bottom}: Task-specific image processing with multiple specialized models.}
\label{fig:intro_figure}
\end{figure}

In recent years, deep learning-based models trained for a wide range of visual perception tasks have enabled the digitization of applications, that previously required manual intervention. These models are typically developed through a rigorous process involving the collection of large datasets, manual annotations, and supervised training of neural networks. However, the need for extensive image data and labor-intensive annotations remains a significant challenge.

A variety of research areas have addressed these challenges, including transfer learning~\cite{mensink2021factors,mustafa2021supervised}, few-shot learning~\cite{sung2018learning,siam2019amp}, zero-shot learning~\cite{xian2018feature}, semi-supervised learning~\cite{reiss2022graph,reiss2021every}, weakly supervised learning~\cite{reiss2023decoupled,li2022decoupled}, and self-supervised learning~\cite{baranchuk2021label,assran2022masked}. Recently, a new approach has emerged: in-context learning~\cite{bar2022visual,wang2023images,bai2023sequential}. In this paradigm, a model is trained to derive the correct output  $\mathcal{O}$ for a query image $\mathcal{Q}$ given a task-specific context $\mathcal{C}$. This approach is akin to few-shot learning but extends to more diverse tasks, combining few-shot and multi-task learning.

In this work, we investigate whether visual in-context learning can be adapted to handle not only individual tasks but also entire task sequences. The ability to instruct a model to process images through a structured sequence of tasks at test time -- such as denoising, artifact removal, and target structure segmentation -- would enable users, \eg, medical doctors, to define highly individualized image processing pipelines without expensive re-training of a specialized model for each subtask.
This flexibility would allow for the dynamic redefinition of image processing pipelines using a single model while upholding transparency, as step-by-step intermediate results can be inspected (\Cref{fig:intro_figure}).
Our exploration addresses several fundamental questions:
\begin{itemize}
\item \textbf{Codebook Limitations (\cref{sec:codebook}):} We examine the limitations of codebooks in visual in-context learning and propose improvements to better capture diverse task outputs.
\item \textbf{Dataset Enrichment (\cref{sec:synth_task_enrich} \& \cref{sec:comp_task_sampling}):} We present a method to enrich datasets with segmentation labels, creating informative task sequences for in-context training.
\item \textbf{Training Objectives (\cref{sec:training_incontext}):} By training codebooks and in-context learners with diverse data and objectives, we gain insight into what works and identify areas needing advancements for compositional medical tasks.
\end{itemize}

Our findings contribute to the development of more flexible and efficient visual in-context learning models capable of handling complex, multi-step tasks in medical imaging.

\section{Related work}
\label{sec:relwork}

\begin{figure}
    \centering
    \begin{subfigure}[b]{0.5\linewidth}
        \includegraphics[height=0.325\linewidth,page=2]{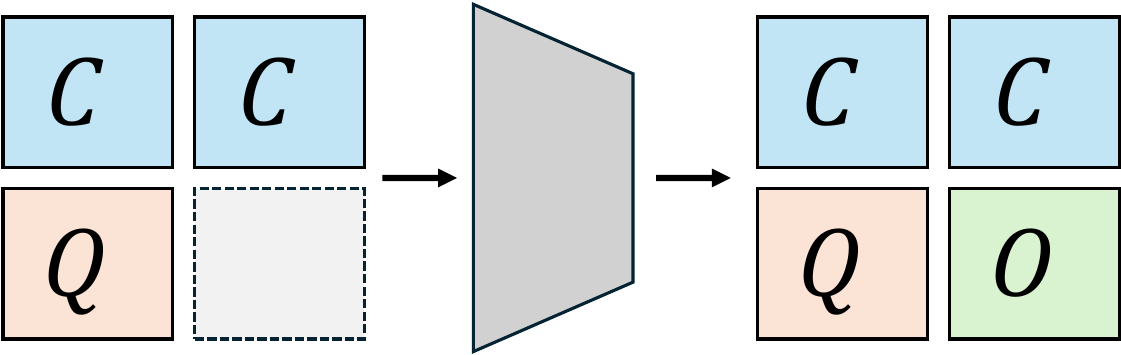}
        \caption{Inpainting with imagegrids.}
        \label{fig:first}
    \end{subfigure}$\phantom{00}$\begin{subfigure}[b]{0.5\linewidth}    
        \includegraphics[height=0.37\linewidth,page=3]{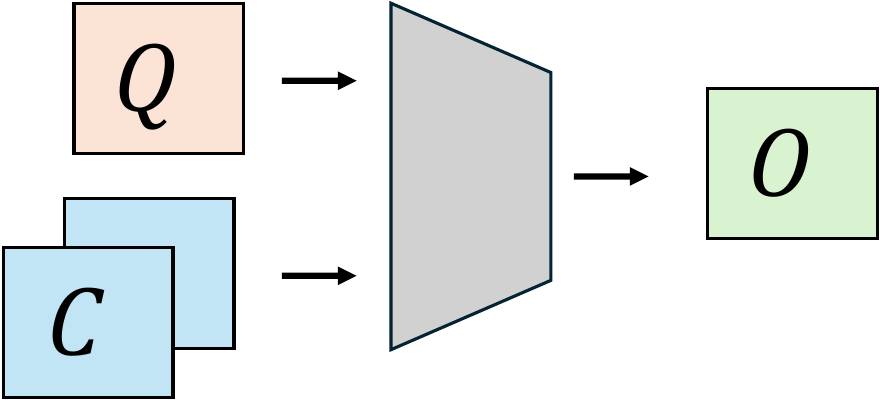}
        \caption{Explicit conditioning.}
        \label{fig:second}    
    \end{subfigure} 
    \begin{subfigure}[b]{\linewidth}    
        \includegraphics[width=\linewidth,page=4]{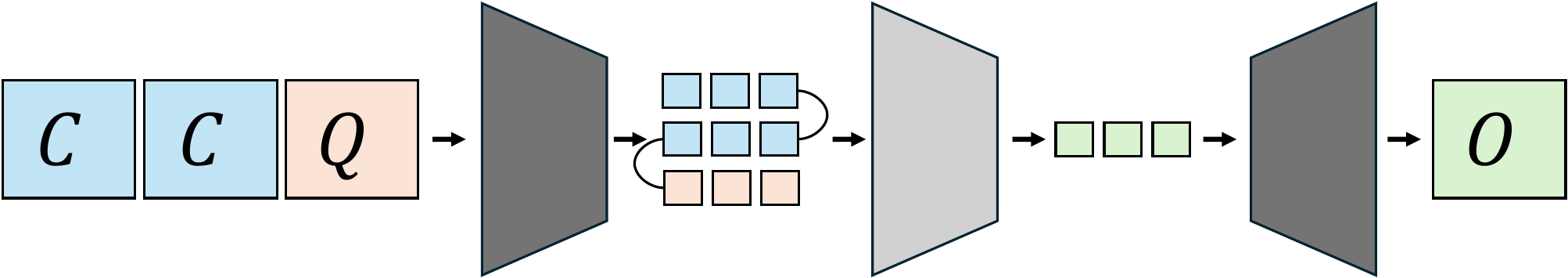}
        \caption{Sequential modeling.}
        \label{fig:third}
    \end{subfigure} 
    \begin{subfigure}[b]{\linewidth}    
        \includegraphics[width=\linewidth,page=5]{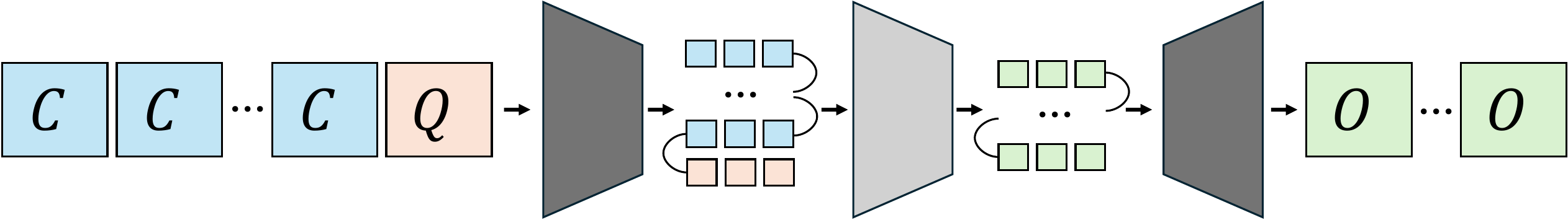}
        \caption{Composite tasks as sequences: Context encodes sequential tasks, the generated output is of sequential nature as well.}
        \label{fig:fourth}    
    \end{subfigure} 
    \caption{Visual in-context learning architectures and paradigms.}
\label{fig:vicl_paradimgs}
\end{figure}

In literature, different choices on how to design visual in-context learners are proposed.
One branch of approaches~\cite{bar2022visual,wang2023images,wang2023seggpt,ren2024medical} encodes the set of reference images associated with annotations directly as a single input image by forming a grid with them (see~\Cref{fig:first}).
The task to be addressed is described in a contextual image-task pair $\mathcal{C}$ in the first row, while in the last row, the query image $\mathcal{Q}$ for which the task specified must be solved is inserted.
The corresponding position of the output to $\mathcal{Q}$ is left blank.
The visual in-context learning model is then trained to inpaint the missing output image $\mathcal{O}$, thereby enabling prompting with new contexts $\mathcal{C}$ to solve new tasks for new queries.

In the medical domain, a different design of the in-context learning process has been proposed~\cite{czolbe2023neuralizer, butoi2023universeg}.
There, networks encode each image-task pair in the context set together with the representation of the query image pairwise and aggregate the information between all image-task pair-query combinations with a pooling operation, enabling arbitrarily many context pairs to be used (see~\Cref{fig:second}).

The third design choice for visual in-context learning frames the setting as a masked- or next token prediction task on input sequences~\cite{bai2023sequential,lu2022unified,guo2024dataefficient} as common in natural language processing~\cite{radford2019language,devlin2018bert}.
To obtain sequences of discrete tokens, the context image-task pairs and query image are arranged into a sequence and processed to discrete tokens via a codebook, more explicitly, the quantization layer~\cite{van2017neural} of a pre-trained VQ-GAN~\cite{esser2021taming,chang2022maskgit,chang2023muse}.
The masked token prediction is done in the latent space, spanned by the codebook (see~\Cref{fig:third}).
To get the in-context prediction for a given query image $\mathcal{Q}$ in inference, the learner is given the contextual image-task pairs $\mathcal{C}$ as token sequence, then the tokens of the query image $\mathcal{Q}$ are appended, and through the training objective of reconstructing masked portions of the sequence, the model predicts tokens corresponding to the output, conditioned on $\mathcal{CQ}$.
Finally, the output tokens can be decoded back into the image space with the codebook.

Further related studies include the design of visual prompting strategies~\cite{zhang2023makes,sun2023exploring}, making image tokenization more concise~\cite{yu2024image}, addressing in-context segmentation ~\cite{wang2023seggpt,butoi2023universeg,rakic2024tyche}, in-context learning on video~\cite{zhang2024video} and visual in-context learning with diffusion models~\cite{gu2024analogist,yang2024imagebrush}.

In contrast to previous approaches, we explore visual in-context learning on compositional tasks, \ie, our context and our outputs encompass sequences of tasks.
To do this, we build on top of the sequential modeling paradigm and extend it to such compositional tasks (see~\Cref{fig:fourth}).

\section{Preliminaries}

\subsection{Compositional in-context learning}
\label{sec:prelim_comp}
We define the task of in-context learning as training a model $\theta(\cdot)$ which, based on a context $\mathcal{C}$ and a query input $\mathcal{Q}$ predicts the contextual output $\mathcal{O}$ for the query.
In visual in-context learning, all constituents in $\mathcal{C}, \mathcal{Q}, \mathcal{O}$ are images of shape $\mathbb{R}^{3 \times W \times H}$.
The context $\mathcal{C}$ contains a task specification, \ie, a small set of images and corresponding task-specific annotations $\mathcal{C} = \{c_0, ..., c_n\}$, that exemplify input-output relations, where $n$ is the number of example input-output relations that specify the task to be solved.
As such, the model needs to capture these relations and apply the task-specific transformation from the query to the output: $\theta(\mathcal{C}, \mathcal{Q}) = \mathcal{O}$.
The model is trained to generalize to novel contexts $\mathcal{C}$, which encode new tasks at test time.

We investigate a specific type of visual in-context learning, where the constituents $c_i$ of the context set are not pairs of image and annotation, but a long sequence of interdependent tasks $c_i \in \mathbb{R}^{t \times 3 \times W \times H}$, where $t$ indicates the number of tasks in the composite task.
This adaptation also entails changes to $\mathcal{O}$, which previously was a singular output image, but now becomes a sequence $\mathcal{O} \in \mathbb{R}^{(t-1) \times 3 \times W \times H}$.
As such, models need to produce an image sequence of $(t-1)$ intermediate task outputs for a query image $\mathcal{Q}$.

\input{content/tables/codebook_eval}

\subsection{Visual codebooks}
To cope with the high dimensional nature of visual in-context learning on compositional tasks, with input $\mathcal{CQ} \in \mathbb{R}^{(n \cdot t + 1) \times 3 \times W \times H}$ and output $\mathcal{O} \in \mathbb{R}^{(t-1) \times 3 \times W \times H}$, a codebook $\phi(\cdot)$ is an essential tool to reduce the dimensionality.
Often parameterized with an auto-encoding neural network with quantization layer~\cite{van2017neural,esser2021taming}, a codebook transforms an image $x$ into a sequence of $q$ discrete tokens $\phi(x) \in [0, \dots, N - 1]^q$, where $N$ is the vocabulary size, \ie, the number of discrete tokens in the codebook.
With such a codebook processing of $\mathcal{CQ} \in \mathbb{N}^{(n \cdot t + 1) \times q}$ to $\mathcal{O} \in \mathbb{N}^{(t-1) \times q}$ is computationally palpable, as $\phi(\cdot)$ is built to produce $q \ll 3 \times W \times H$.
The reverse direction from discrete token space back into the image space is done with a decoder $\phi^{-1}(\cdot)$, both $\phi(\cdot)$ and $ \phi^{-1}(\cdot)$ are trained jointly.

The number of tokens per image $q$ in common codebooks, \eg, VQ-GAN~\cite{esser2021taming} codebooks, is influenced by the image size and the network architecture, as spatially smaller latent representations lead to fewer tokens.

\section{Analysis of task recovery in codebooks}
\label{sec:codebook}
Codebooks are an enabling factor for visual in-context learning on compositional tasks, but they also constrain what can be learned when transferring $\mathcal{C}\mathcal{Q}$ to $\mathcal{O}$, as this transfer now happens in token space.
To investigate how much codebooks limit in-context learning, we measure how well they can recover task-specific outputs in $\mathcal{O}$.

\subsection{Codebooks for task-specific reconstruction}

Therefore, we run a small set of preliminary experiments, where we simply take pre-trained codebooks (model zoo of~\cite{rombach2021highresolution,esser2021taming}), map different task outputs into the token space and directly reconstruct them back into the image space to evaluate them with respective task metrics.
In~\Cref{fig:codebook_spiderplot}, we consider the tasks image reconstruction, image denoising, and semantic segmentation and determine the upper bounds for any in-context model working with these codebooks.


\textbf{Image reconstruction}:
The codebooks were all trained for the task of image synthesis, a very similar task to this is pixel-wise image reconstruction.
When we look at the capabilities of accurately reconstructing images in~\Cref{fig:spiderplot1}, which can be measured in mean absolute error (MAE), we see, that the best models consistently are under $0.05$ MAE, an acceptable bound for this metric.
This outcome is not too surprising, as the objective function of all the trained codebooks involved a regression loss term, to lead to small color-value deviations~\cite{esser2021taming}.
What's more interesting is, that the errors of the best models stay low for non-natural image domains, on which the models were not specifically trained, such as optical coherence tomography (cirrus, spectralis, topcon~\cite{bogunovic2019retouch}), electron microscopy (hela, jurkat, macrophage~\cite{heinrich2021whole}) or skin-lesion images (HAM10000~\cite{tschandl2018ham10000}).

\textbf{Image compression and denoising}:
When denoising is the task to be addressed a question that arises is, whether codebooks can even represent image noise, \ie, whether the tokens allow for noise reconstruction.
When increasing noise-intensity in natural images (BSDS500~\cite{MartinFTM01}), codebooks can not recover this induced noise.
In~\Cref{fig:spiderplot2}, more noise leads to a higher Root-mean-square error (RMSE), which is unsurprising, as neither the training data nor the loss functions of the codebooks are tuned for recovering noise and thereby, as a side-product of the compression via auto-encoding, noise is ignored during reconstruction.
For the task of image denoising, the Peak signal-to-noise ratio (PSNR) is a commonly reported metric, which is bound to $27.46$ on BSDS500 when using the best pre-trained codebook to reconstruct the clean data.
As such denoising can not exceed this threshold for this data.


\textbf{Semantic segmentation}:
To evaluate the upper bound for image segmentation, we tokenize segmentation maps of different datasets and reconstruct them.
To compute the Intersection over Union (IoU), we map the reconstructed color values to the closest valid color in the input segmentation map.
We notice, that for datasets where binary segmentation is done, such as for skin lesion segmentation (HAM10000~\cite{tschandl2018ham10000}), the upper IoU bound is almost pixel-perfect, which might be due to the simple topology of this data, \ie, blob-like structures.
When moving from binary- to multi-class segmentation, the upper bound successively degrades to $98\%$ for three classes (cirrus, spectralis, topcon~\cite{bogunovic2019retouch}) and when increasing to $36$ classes (hela-2~\cite{heinrich2021whole}) the upper bound is $71\%$ IoU.
This likely has two causes, first, more classes lead to finer segmentation structures which are harder to reconstruct, and second, more classes lead to more colors in the segmentation maps which in turn leads to a fuzzier assignment to the valid colors.
We exemplify this to the extreme, with experiments on the ATLAS dataset~\cite{jaus2023towards} with over $142$ classes.
There, the codebooks are not capable to precisely reconstruct colors such that they are mapped to the correct classes, leading to an upper IoU of merely $29\%$.
Yet, we notice that if codebooks are prompted with segmentation maps utilizing different color schemes ($v_0 - v_2$) for these $142$ classes, the upper bound changes drastically to an IoU of $44\%$.
As such, codebook-based token spaces for in-context learning are sensitive to visual prompting~\cite{sun2023exploring}.

\textbf{Summary on recovering task information}
Building an in-context learner on top of a codebook that was trained for image synthesis propagates its capabilities, but also its limitations.
Concretely, when the task includes image reconstruction or recovering simple binary structures, near perfect performance may be achieved by the learner using the codebook.
Yet, when the tasks get highly discriminative, \ie, discrimination between a high number of semantic concepts, or when fine-grained structures need to be identified, codebooks pre-trained on natural images produce errors.



\subsection{Task-informed codebooks}
\label{sec:taskinformedcodebooks}
To address this gap in representing task-related outputs, such as segmentation maps, we propose a simple adaptation to codebook training: train the codebook on imaging data \textit{and} on task outputs.
While this data-centric solution for codebooks to better represent task data sounds simple, in practice we experiment with two important aspects.

\textit{Data balancing}: To capture multiple modalities, \ie, image data and different task output types (\eg, segmentation maps, bounding boxes, etc.), we explore how to best balance the training data.
One aspect is the balance between different task output types, the second aspect is to balance sampling from different datasets when multiple are in use.

\textit{Color remapping augmentation}: We noticed in our preliminary investigation on codebooks, that different color schemes can have a major effect on the upper performance bound under a given codebook.
To overcome this, we utilize the idea of augmenting the semantic task outputs.
During training of the codebook, we resample the colors in segmentation maps, bounding boxes, etc. to random different colors.
This ensures, that the codebook is not only capable to represent a specific color map, but can generalize to prompts with new class-color definitions.

With a codebook that tokenizes images and task outputs, we move to in-context learning on compositional tasks next.

\section{Compositional visual in-context learning}
\label{sec:enrich}
\begin{figure*}[t]
    \centering
    \includegraphics[width=\textwidth,page=5]{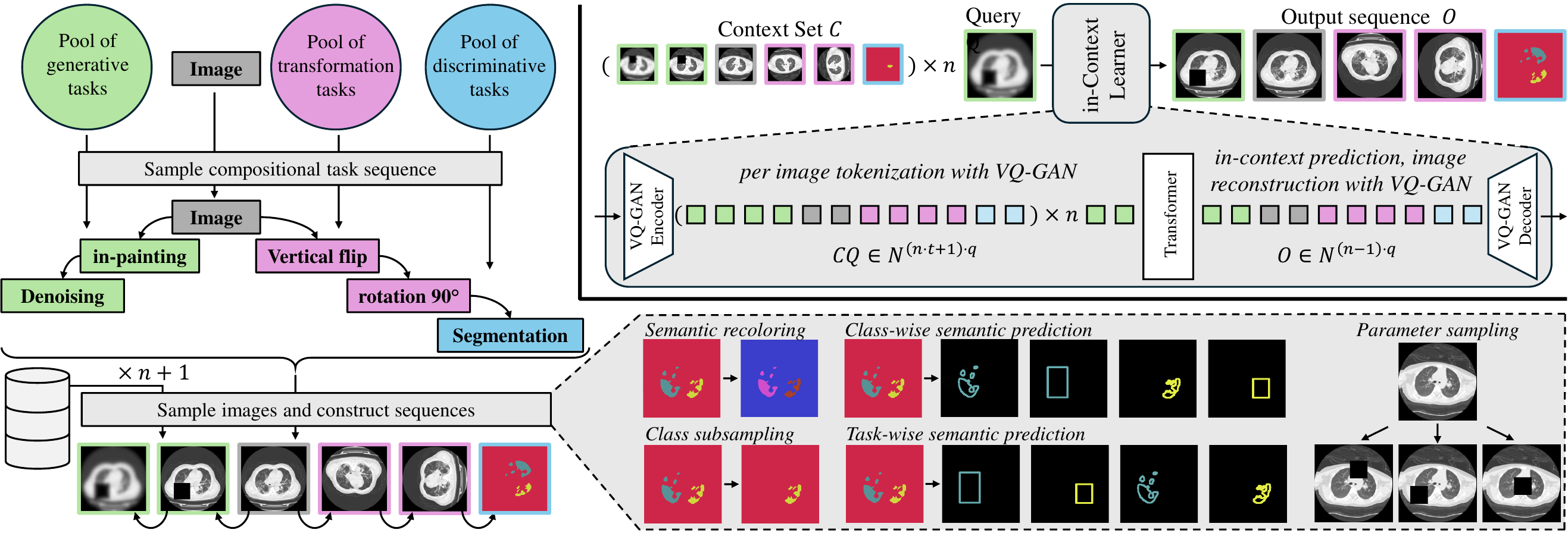}
    \caption{\underline{On the left}, we present the sequence structure, which we enforce to construct compositional tasks, with some guardrails shown on the \underline{bottom right}. \underline{The top right} shows the general processing pipeline of compositional tasks through a visual in-context learner.}
    \label{fig:method}
\end{figure*}

First, we obtain training data for compositional tasks via synthetic data generation, then we outline pre-training objectives for visual in-context learning.
In~\Cref{fig:method}, we show our compositional visual in-context learning pipeline.

\subsection{Synthetic task enrichment}
\label{sec:synth_task_enrich}
To advance towards the setting of compositional in-context learning, the basic requirement is access to compositional task sequences $\mathcal{C}\mathcal{Q}\mathcal{O}$ to train with.
As such, we start with describing a data-centric pipeline to obtain multiple vision tasks from pre-existing semantic segmentation datasets.
In order to generalize to new task sequences at test time, the training task sequences need to be highly diverse and cover different tasks.
We first enrich the dataset by either utilizing the images of the dataset to setup additional generative- and transformation tasks or the segmentation annotation to derive annotation variants for additional discriminative tasks.

\textbf{Generative Tasks:} The first type of tasks we formulate are generative tasks. Here the task is to restore the original image from a degraded version.
For example, we lower the resolution of an image which serves as input and take the original image as prediction target to form the task of \textit{super resolution}.
Similarly, we formulate an \textit{inpainting} task, and \textit{denoising} task by either cutting out portions of the original image or adding Gaussian noise.
Lastly, for other types of degradations or photometric adaptations to form input-output task pairs we add \textit{color jitter} to remove, \textit{invert} the intensity values of the original image or alter its \textit{brightness}.

\textbf{Geometric transformation tasks:} In addition to altering the image content with artifacts and degradation and successively reconstructing the obstructed information, we formulate transformation tasks.
These tasks are based on the geometric transformations: \textit{horizontal flip}, \textit{vertical flip} and \textit{rotation} by $\{90^\circ, 180^\circ, 270^\circ\}$.
Here, the input is the original image and the output is the geometrically transformed image, resulting in tasks that require global information exchange when formulated as image-to-image tasks.

\textbf{Discriminative tasks:} The last group of tasks we derive from the original dataset are based on segmentation annotations.
We extract the edges of all semantic classes in the annotations to get \textit{semantic edge} annotations.
Akin to object detection, we obtain detection targets by drawing \textit{semantic boxes} around individual segments in the images.
Lastly, we simulate two exotic annotation types for more diversity, so called \textit{skeletons} (medial axes) of segments which are thin lines describing their topology as well as central \textit{points} on segments.
Skeletons and points are related to scribbles~\cite{marinov2024deep} from interactive- or weakly-supervised learning~\cite{reiss2023decoupled}.

The extracted generative-, transformation- and discriminative task information will form the basis of our compositional task sequences.
As in-context learning has been mainly done with large heaps of data~\cite{bai2023sequential,guo2024dataefficient,DBLP:journals/corr/abs-2005-14165}, this enrichment is done for $38$ medical segmentation datasets~\cite{ma2024segment}.

\subsection{Compositional task sampling}
\label{sec:comp_task_sampling}
With the enriched dataset comprised of diverse vision tasks, we are equipped to build task sequences.
Crucially, when sampling task sequences from the dataset, all $n + 1$ task sequences in $\mathcal{C}\mathcal{Q}\mathcal{O}$, \ie, all $c_i$ and the concatenation $\mathcal{Q}\mathcal{O}$ need to obey the same task-sequence logic.
While randomly drawing tasks, putting them into a sequence and based on it drawing samples from the dataset successively leads to the highest diversity in task sequences, this approach fails in reality.
This is due to task sequences becoming extremely poor in information.
For discriminative tasks, this is due to the fact, that randomly drawing images often leads to sequences that do not share the same classes, and as such $\mathcal{O}$ may contain classes that are not present in any $c_i$, and thereby the task to predict them from $\mathcal{C}$ is ill-posed.
Similarly, for generative tasks, sequencing is key, as randomly sampling an image into the sequence and thereafter its low resolution version from the \textit{super resolution} task amounts to a rather simple task sequence encoding down scaling.

\subsubsection{Controlling task sequence generation}
With this insight, we formulate guardrails within which task sequences can be sampled randomly, to balance the information content of the sequences with their diversity.

We enforce a task sequence structure that follows: \textit{$\{$\generative{generative}$\}$ image $\{$\transformation{transformation}$\}$ $\{$\discriminative{discriminative}$\}$}.
For each task family in brackets, multiple task instances may be sampled, \eg, an explicit task sequence of the structure \textit{\generative{inpaint}}$\to$\textit{\generative{denoise}}$\to$\textit{image}$\to$\textit{\transformation{rotate $90^\circ$}}$\to$\textit{\discriminative{segmentation}} would be valid.
With this structure, task sequences can be computed in a straightforward fashion.
First, generative tasks, which obfuscate the original image $x$ are applied from right to left $\varphi_{\text{\generative{inpaint}}}(\varphi_{\text{\generative{denoise}}}(x)),\varphi_{\text{\generative{denoise}}}(x), x$, then geometric transformations are applied and added to the right of the sequence: $\varphi_{\generative{\text{inpaint}}}(\varphi_{\generative{\text{denoise}}}(x)),\varphi_{\generative{\text{denoise}}}(x), x, \varphi_{\transformation{$\text{rotate}~90^\circ$}}(x)$ and finally the discriminative tasks are added, \ie, the semantic annotation $x^\discriminative{\text{seg}}$, which has to be transformed with all geometric transformations first:
$\varphi_{\generative{\text{inpaint}}}(\varphi_{\generative{\text{denoise}}}(x)),\varphi_{\generative{\text{denoise}}}(x), x,$ $\varphi_{\transformation{$\text{rotate}~90^\circ$}}(x), \varphi_{\transformation{$\text{rotate}~90^\circ$}}(x^\discriminative{\text{seg}})$.
We directly see, that some tasks are dependent on others, \eg, the segmentation task operates on the rotated version of $x$, or, inpainting operates on the noised image.
This highlights the interdependence and compositional nature of our synthetic task sequences.

On top of this task structure, we add a few more rules and guardrails for information-rich task sequences.
For discriminative tasks, we increase the diversity by sub-sampling the available classes, such that not all classes will be present in all task sequences, but consecutive task sequence sampling may contain different class-subsets.
This adds an incentive to adhere to the classes in the context $\mathcal{C}$ and only produce outputs $\mathcal{O}$ that contain the same class subset.
This guardrail is designed to counteract a degradation towards always predicting all classes associated to a given image.

To further increase diversity, we either present the discriminative task output for all classes in one image or adapt the discriminative task to a class-wise task sequence, where the task output is predicted for each class individually.
In case a task sequence contains multiple discriminative tasks, the ordering can either be on a task basis, \ie, different tasks in sequence, or on a class basis, \ie, the task outputs are grouped by class, meaning all tasks are solved for one class before the next class is handled.

To ensure flexibility, we recolor the semantic annotations consistently in $\mathcal{CQO}$ and assign a new color for each class randomly to anticipate new class-colors that may occur in testing.
Further, when sampling the context set $\mathcal{C}$ we make sure that it contains all classes present in $\mathcal{O}$. 

For more diversity in the different tasks, we sample their parameters, such as mean and variance for Gaussian noise, the cut-out positions for inpainting and the downscale factor for super resolution, line width of semantic edges, boxes or skeletons, and rotation angles.
All generation guardrails are visualized on the bottom right of~\Cref{fig:method}.

\subsubsection{Controlling task sequence length}
A consideration when computing task sequences is the sequence length, as too long sequences make learning correlations between sequence elements difficult and computationally expensive.
For adequate costs, we bound the total length $|\mathcal{CQO}|$, in our case to $30$ images, which limits the number of tasks in a sequence to $15$ (as the minimal compositional task is $|c_0\mathcal{QO}| = 30$, and thus $|c_0|, |\mathcal{QO}| = 15$).

Note, longer sequences are possible, yet, a sequence of $30$ images equates to a sequence length of $30 \cdot q$ in token space.
For common codebooks $q$ ranges between $144$ and $4,096$ tokens per image, such long sequences require extensive compute infrastructure and may require model architectures tailored for excessive sequence lengths~\cite{gu2023mamba,beck2024xlstm}.

\subsection{Training compositional in-context learners}
\label{sec:training_incontext}
In our work, we mainly investigate \textit{how} to enable neural networks to learn compositional vision tasks in context.
The question which network architecture might be best suited for this setting is beyond this investigation.
Thus, we resort to the tried and tested Transformer architecture~\cite{DBLP:conf/nips/VaswaniSPUJGKP17,esser2021taming,radford2019language}, which we utilize as in-context learner.

The more fundamental question that needs to be answered is with which optimization objective to train the transformer.
For this, we orient ourselves at masking- and subsequent reconstruction-based training strategies which have been successfully used in language modeling~\cite{DBLP:journals/corr/abs-1810-04805,DBLP:journals/corr/abs-2005-14165,DBLP:journals/corr/abs-1910-01108}, image pre-training~\cite{assran2022masked,he2022masked,assran2023self} and visual in-context learning~\cite{bai2023sequential,guo2024dataefficient}.
As such, we transfer the compositional task sequences $\mathcal{CQO}$ into token space to form a long token sequence.
Then we randomly switch out tokens in this sequence with random tokens, whereas the transformer is tasked with identifying those malicious tokens and predicting the correct ones.
Specifically, the manipulated token sequence serves as input to the model which predicts a softmax output over all codebook entries for each sequence element.
At each location of a malicious (\ie, masked) token the cross-entropy loss between the prediction and the ground-truth sequence $\mathcal{CQO}$ is back propagated.

\begin{figure}
    \centering
    \includegraphics[width=\linewidth,page=6]{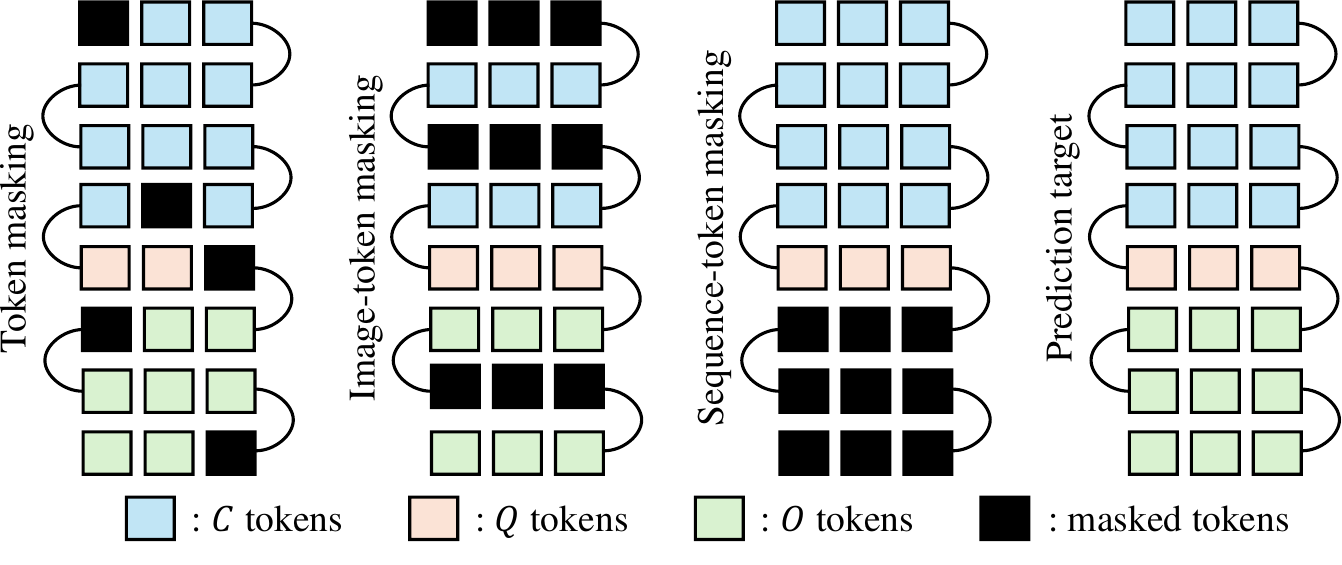}
    \caption{Token masking variants for training visual in-context learners, each row of tokens symbolizes the tokens of one image.}
    \label{fig:masking}
\end{figure}

The concrete masking strategy to best train the transformer for recovering the contextual output $\mathcal{O}$ at test time is not clear, three strategies are shown in~\Cref{fig:masking}.
Random \textit{token masking} is one possible strategy, where in the sequence a portion $p$ of tokens are switched out randomly as done in language modeling and vision transformer pre-training~\cite{DBLP:journals/corr/abs-2005-14165,he2022masked}.
Adapting the idea of masking out entire words in language modeling to our setting aligns better with switching out all tokens that encode individual images, which is what we refer to as \textit{image-token masking}.
Lastly, as we are able to produce the composite tasks of context set $\mathcal{C}$, query $\mathcal{Q}$ and output $\mathcal{O}$, we can, in training, mirror the task of compositional in-context prediction by simply switching out all tokens in $\mathcal{O}$, which we term \textit{sequence-token masking}.

\textbf{Special tokens}
To inform the transformer about the end of tokens that correspond to one image, we add $k$ special tokens $\langle$i-END$\rangle$ after the tokens of each image into the sequence.
Similarly, we add $k$ $\langle$c-END$\rangle$ tokens, when a context element $c_i$ is complete.
We add multiple such tokens, as transformers have been shown to benefit from additional special tokens to store information~\cite{darcet2023vision}.





{
\addtolength{\tabcolsep}{-5.1px}
\begin{table*}[t]
    \scriptsize
    \centering
    \begin{tabular}{l|ccc|ccccccc|ccc|ccccc}
         \toprule
         Method & \multicolumn{3}{c}{Masking} & \multicolumn{7}{c}{\generative{Generative}} & \multicolumn{3}{c}{\transformation{Transformation}} & \multicolumn{5}{c}{\discriminative{Discriminative}} \\
          & $p$ & \#im & $\mathcal{O}\phantom{|}$ &  Super res. & Inpaint & Denoise & Jitter & Invert & Equal. & Bright & Flip (h) & Flip (v) & Rotate & Seg. & Boxes & Points & Skeletons & Edges \\
         & & & & PSNR $\uparrow$ & PSNR $\uparrow$ & PSNR $\uparrow$ & PSNR $\uparrow$ & MSE $\downarrow$ & MSE $\downarrow$ & MSE $\downarrow$ & MSE $\downarrow$ & MSE $\downarrow$ & MSE $\downarrow$ & IoU $\uparrow$ & F-1 $\uparrow$ & F-1 $\uparrow$ &  F-1 $\uparrow$ &  F-1 $\uparrow$ \\
         \midrule
         Copy baseline & -- & -- & -- & $11.477$ & $12.347$ & $11.814$ & $11.585$ & $0.0845$ & $0.0728$ & $0.0864$ & $0.0789$ & $0.0772$ & $0.0802$ & $50.79\%$ & $63.49\%$ & $51.52\%$ & $47.04\%$ & $48.59\%$ \\
         \midrule
         Token masking & $15 \%$ & -- & -- & $9.398$ & $9.948$ & $9.692$ & $9.651$ & $0.1341$ & $0.1310$ & $0.1296$ & $0.1456$ & $0.1484$ & $0.1566$ & $26.49\%$ & $38.17\%$ & $36.54\%$ & $36.49\%$ & $35.32\%$ \\
         Image-token masking & --  & $5$ & -- & $16.063$ & $15.288$ & $16.851$ & $15.726$ & $0.0387$ & $0.0468$ & $0.0425$ & $0.0762$ & $0.0839$ & $0.0943$ & $44.83\%$ & $61.01\%$ & $50.36\%$ & $46.17\%$ & $47.42\%$ \\
         Sequence-token masking & --  & -- & \checkmark & $18.265$ & $19.116$ & $20.131$ & $19.206$ & \underline{$0.0173$} & $0.0167$ & $0.0174$ & $0.0203$ & $0.0200$ & $0.0219$ & $52.60\%$ & $62.26\%$ & $52.79\%$ & \underline{$47.38\%$} & $48.16\%$ \\
         $\phantom{000}\llcorner$ $\langle$END$\rangle$ $k=2$ & --  & -- & \checkmark & \underline{$18.411$} & \underline{$19.282$} & \underline{$20.169$} & \underline{$19.280$} & $0.0176$ & \underline{$0.0162$} & \underline{$0.0171$} & $0.0196$ & \underline{$0.0193$} & $0.0207$ & \underline{$52.82\%$} & $62.32\%$ & $52.68\%$ & $47.17\%$ & $48.27\%$ \\
         $\phantom{000}\llcorner$ $\langle$END$\rangle$ $k=4$ & --  & -- & \checkmark & $18.290$ & $19.226$ & $20.130$ & $19.182$ & $0.0181$ & $0.0163$ & $0.0175$ & \underline{$0.0195$} & $0.0194$ & \underline{$0.0206$} & $52.78\%$ & \underline{$62.36\%$} & \underline{$52.84\%$} & $47.16\%$ & \underline{$48.30\%$} \\
         Mixed masking & $15 \%$ & $3$ & \checkmark & $17.870$ & $18.431$ & $19.393$ & $18.554$ & $0.0211$ & $0.0200$ & $0.0197$ & $0.0236$ & $0.0230$ & $0.0252$ & $50.39\%$ & $62.06\%$ & $52.68\%$ & $46.77\%$ & $47.82\%$ \\
         \midrule
         Longer training ($k=2$)& --  & -- & \checkmark & $\textbf{18.513}$ & $\textbf{19.652}$ & $\textbf{20.531}$ & $\textbf{19.711}$ & $\textbf{0.0165}$ & $\textbf{0.0147}$ & $\textbf{0.0158}$ & $\textbf{0.0165}$ & $\textbf{0.0166}$ & $\textbf{0.0169}$ & $\textbf{61.21\%}$ & $\textbf{65.92\%}$ & $\textbf{55.66\%}$ & $\textbf{49.55\%}$ & $\textbf{50.85\%}$ \\ 

         \midrule

         Upper Codebook bound & -- & -- & -- & $20.354$ & $21.975$ & $22.527$ & $21.800$ & $0.0107$ & $0.0077$ & $0.0109$ & $0.0097$ & $0.0096$ & $0.0098$ & $86.72\%$ & $85.33\%$ & $74.69\%$ & $60.21\%$ & $66.32\%$ \\
         \bottomrule
    \end{tabular}
    \caption{Evaluation of Visual in-Context Learning on compositional medical tasks under different masking pre-training strategies.}
    \label{tab:multi-masking}
\end{table*}
}

{
\addtolength{\tabcolsep}{-0.4em}
\begin{table}[h]
    \scriptsize
    \centering
    \begin{tabular}{lccccc}
         \toprule
         Method & Segment. & Edge detection & Skeletons & Bound. box & Reconst.\\
         & IoU $\uparrow$ &F-1 Score $\uparrow$& F-1 Score $\uparrow$ & F-1 Score $\uparrow$ & MAE $\downarrow$\\
         \midrule
         \multicolumn{6}{c}{\textit{Fixed color semantics}} \\
         VQ-GAN & $59.89 \%$ & $59.53 \%$ & $55.74 \%$ & $60.65 \%$ & $0.09121$ \\
         $\phantom{0}\llcorner$ task training & $ 86.19 \%$ & $ \textbf{95.54} \%$ & $\textbf{96.24} \%$ & $99.45 \%$ & $0.03756$ \\
         $\phantom{00}\llcorner$ task balance & $ \textbf{88.07} \%$ & $ \textbf{95.54} \%$ & $96.08 \%$ & $\textbf{99.74} \%$ & $0.03745$ \\
         $\phantom{000}\llcorner$ recoloring & $ 84.07\%$ & $ 81.23 \%$ & $84.27 \%$ & $84.97 \%$ & $\textbf{0.03681}$ \\
         $\phantom{00}\llcorner$ task + data b. & $ 86.76 \%$ & $ 95.18 \%$ & $96.06 \%$ & $99.59 \%$ & $0.03706$ \\
         $\phantom{000}\llcorner$ recoloring & $ 83.59\%$ & $ 85.04 \%$ & $87.01 \%$ & $91.58 \%$ & $0.03742$ \\
         \multicolumn{6}{c}{\textit{Random color semantics}} \\
         VQ-GAN & $59.12 \%$ & $61.14 \%$ & $56.62 \%$ & $62.51 \%$ & $0.09121$ \\
         $\phantom{0}\llcorner$ task training & $ 66.29 \%$ & $ 72.78 \%$ & $70.06 \%$ & $73.28 \%$ & $0.03756$ \\
         $\phantom{00}\llcorner$ task balance & $ 69.21 \%$ & $ 74.65\%$ & $71.71 \%$ & $74.58 \%$ & $0.03745$ \\
         $\phantom{000}\llcorner$ recoloring & $ 83.00 \%$ & $ 91.45\%$ & $91.19 \%$ & $95.86 \%$ & $\textbf{0.03681}$ \\
         $\phantom{00}\llcorner$ task + data b. & $ 71.15 \%$ & $ 75.65 \%$ & $72.73 \%$ & $75.97 \%$ & $0.03706$ \\
         $\phantom{000}\llcorner$ recoloring & $ \textbf{83.19} \%$ & $ \textbf{91.93}\%$ & $\textbf{92.15} \%$ & $\textbf{97.46} \%$ & $0.03742$ \\
         \bottomrule
    \end{tabular}
    \caption{Upper performance bounds set by trained codebooks when encoding and reconstructing images and task outputs.}
    \label{tab:codebook_training_eval}
\end{table}
}

\section{Experiments}
\label{sec:experiments}
In this section, we outline the training and evaluation of visual in-context learners for compositional medical tasks.

\subsection{Datasets \& evaluation setting}
We utilize $38$ datasets from the MedSAM dataset collection~\cite{ma2024segment}, which is comprised of diverse medical datasets, covering different imaging modalities such as computed tomography, optical coherence tomography, x-ray, ultrasound and images of skin lesions.
We extract images and segmentation maps, encode them as RGB images, rescale them to $200 \times 200$ and enrich them as outlined in~\Cref{sec:synth_task_enrich}.

Training of codebooks is done on individual images and task outputs from our enriched MedSAM data.
The codebook in turn is used to tokenize compositional tasks $\mathcal{CQO}$ on an image-wise basis.
Our codebook processes $200 \times 200$ images into $q = 144$ tokens, with at most $30$ images in task sequences the max length is $4,320$ (without special tokens).

The visual in-context learners are trained on compositional tasks as generated with our sampling process in~\Cref{sec:comp_task_sampling} and tokenization with the trained codebook.

We evaluate the efficacy of the codebook to reconstruct task-specific outputs and the performance of the visual in-context learner to predict task sequences with the metrics Intersection over Union (IoU), F-1 Score, Mean Average Error (MAE), Root-Mean-Square Error (RMSE) and Peak Signal-to-Noise Ratio (PSNR), depending on the task.

\subsection{Implementation details}
The visual codebooks we train are VQ-GANs~\cite{esser2021taming} with $70$M parameters that are pre-trained on ImageNet~\cite{deng2009imagenet}.
The visual in-context learners that we train are GPT2 transformers~\cite{radford2019language} as implemented in~\cite{esser2021taming} with $195$M parameters.

\textit{Codebook training}:
Our codebook is the VQ-GAN $f=16, KL (d=16)$ variant\footnote{\url{https://github.com/CompVis/latent-diffusion/}}, the amount of tokens to be learned is $N = 16,384$.
We fine-tune it on the enriched MedSAM data, while we omit the discriminator loss of the VQ-GAN setup, as we noticed diverging results when utilizing it.
Training is done for one epoch with a batch size of $96$ and a total of $5$M images and task outputs combined.

\textit{In-context learner training}:
The GPT2 transformer is optimized $8$ epochs on $222$K synthetically generated compositional tasks.
Due to memory constraints, we train with a batch size of four, a context window of $4,500$ and operate on $16,386$ tokens, \ie, codebook size plus two special tokens, our default number of inserted special tokens is $k=1$.
All models are trained on $4\times$NVIDIA A100 with 40GB.

\begin{figure*}[t]
    \centering
    \includegraphics[width=\textwidth, page=2]{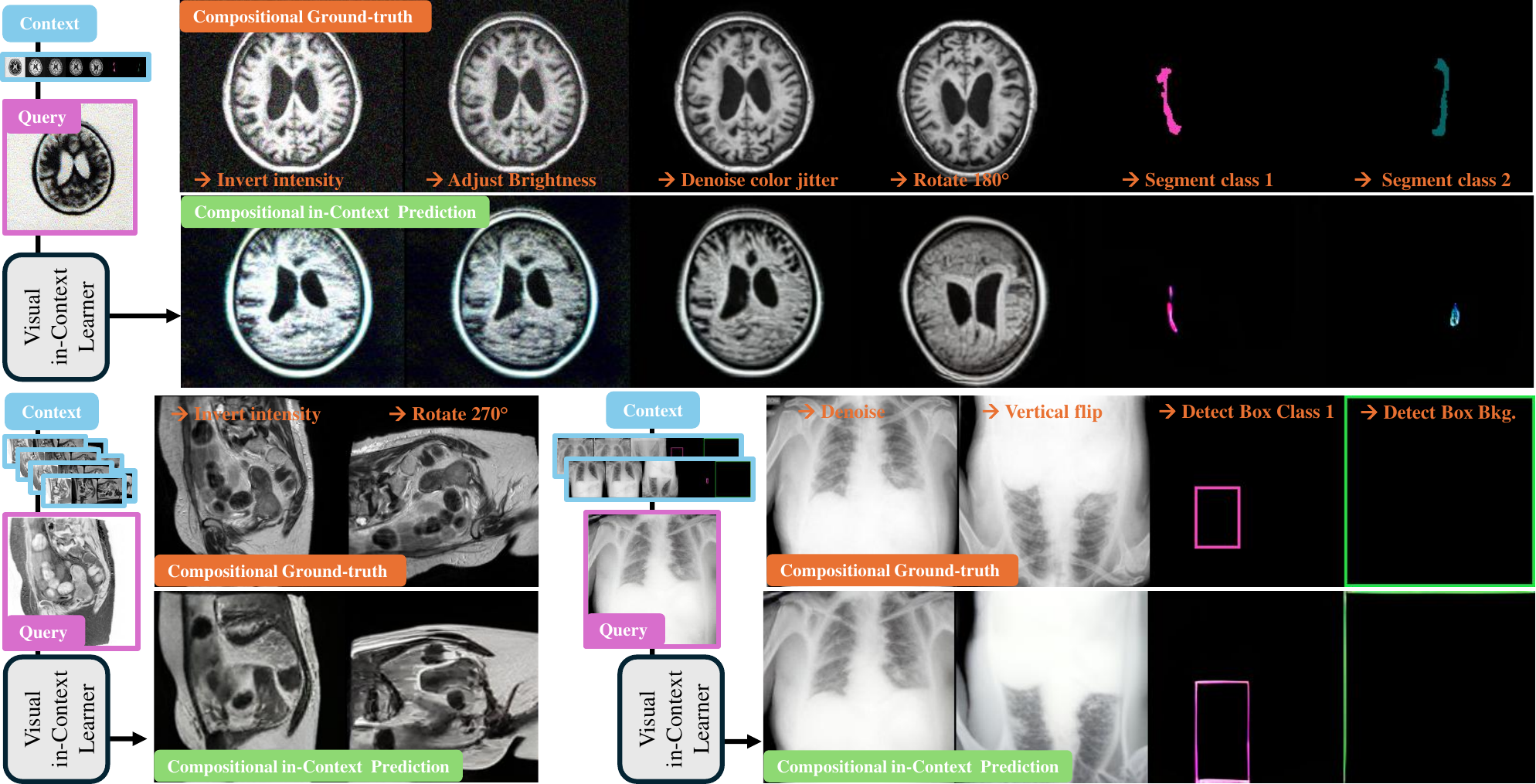}
    \caption{We show three different qualitative compositional medical task predictions by the best visual in-context learning variant.}
    \label{fig:qualitative_testtasks}
\end{figure*}

\subsection{Quantitative codebook evaluation}
In~\Cref{tab:codebook_training_eval}, we show how well different codebooks can reconstruct MedSAM images and importantly the \textit{discriminative task} outputs.
First, we evaluate a pre-trained VQ-GAN codebook in row one, where we see that it struggles severely to represent \textit{discriminative task} related information in MedSAM with scores between $55.74-60.65\%$, merely the image reconstruction error is low.
When fine-tuning it on MedSAM task data, we see that for \textit{discriminative tasks} the results improve drastically to scores ranging between $86.19 -99.45\%$ (row two).
Balancing the training data to contain each task output and image to a similar extent leads to slightly higher upper codebook bounds for segmentation and boxes.
Further, when balancing the data with respect to both tasks and oversampling small datasets in MedSAM for all datasets to be represented equally, we observe similar results (row five).
In the first six rows, we evaluate on task outputs with fixed color schemes, we can see, that in this evaluation scenario adding the random color augmentation (rows four and six) from~\Cref{sec:taskinformedcodebooks} lowers performance.

When switching to a more realistic setup, where \textit{discriminative tasks} come with arbitrary colors during testing (rows seven to twelve), we see the ranking shift.
The results of all codebooks trained without color remapping augmentation diminish severely on new class-colors.
The clear winner is the VQ-GAN with task- and dataset balancing and color remapping augmentation (last row).
For our visual in-context learning experiments, we utilize this codebook.

\subsection{Quantitative results on compositional tasks}
In~\Cref{sec:training_incontext}, we define three different masking strategies for pre-training the visual in-context learner.
We compare them in~\Cref{tab:multi-masking}, where we train the in-context learner for $8$ epochs either with switching out $p=15\%$ of the tokens randomly, or around $15\%$ of image-tokens (\ie, tokens of $5$ full images), or all tokens in the output-sequence $\mathcal{O}$ are switched out.
The results are computed on $2,000$ test compositional tasks formed from image- and task outputs excluded from the training set.
The lower copy baseline, which simply predicts samples from the context set, and the upper codebook bound delimit the results.

The pre-training strategy with simple token masking (row two), yields the worst results when evaluating all intermediate task outputs in the predictions $\mathcal{O}$ for unseen compositional tasks.
Identifying and switching randomly switched tokens to the correct ones may be solved by attending to the local context of a token and does not require learning long range dependencies between contexts $c_i$ and $\mathcal{O}$.
When dropping out all tokens of images, we notice a substantial improvement in metrics (row three), the model starts to draw correlations between context- and output sequences, filling in coherent image-tokens.
Lastly, we mirror the test scenario of switching out all tokens of $\mathcal{O}$ and predicting the correct ones.
This strategy leads to a substantial improvement in all metrics (row four), the network now captures the interdependence between task sequences in the context and patterns to be decoded for predicting the sequence $\mathcal{O}$.
Mixing all strategies above does not improve results (row seven), increasing the number of $\langle\text{END}\rangle$ tokens to $k=2$ (row five) and $k=4$ (row six) yields slight improvements, with $k=2$ coming out on top.
By training the variant with $k=2$ and \textit{sequence-token masking}for $52$ epochs (row eight) the results improve further.
We see \textit{generative-} and \textit{transformation tasks} can be better captured by the visual in-context learners, while they struggle with connecting image patterns with semantics in \textit{discriminative tasks}, especially fine structures such as boxes, points, edges or skeletons.

\subsection{Qualitative results on compositional tasks}
In~\Cref{fig:qualitative_testtasks}, we display qualitative compositional task predictions.
The first sequence prediction (top) indicates that the visual in-context learner can capture the structure of the tasks to be solved coherently, \ie, follow the instructions from the context set.
The bottom two outputs operate on different imaging modalities, all of which are captured well, highlighting the multi-modal capabilities of the model.

\section{Conclusion}
\label{sec:conclusion}
This paper explored visual in-context learning on compositional medical tasks.
We identified prerequisites and opened a pathway to train transformer-based in-context learners on synthetic task sequences.
The trained models are capable of solving complex vision problems step by step, thereby making their response interpretable through verifiable intermediate results -- important in medical imaging.
Such step-by-step processing may in the future enable visual thinking processes as in language modeling~\cite{wu2023thinking,guo2025deepseek}.
Yet, as we only took a first step, challenges remain: codebooks need to represent fine structures better, training objectives need to be improved to better capture image-semantics relations, and capabilities on new task distributions need to be explored.

\noindent\textbf{Acknowledgments} This work was supported by funding from the pilot program Core-Informatics of the Helmholtz Association (HGF) and by the joint research school “HIDSS4Health – Helmholtz Information and Data Science School for Health. The authors gratefully acknowledge the computing time provided on the high-performance computer HoreKa by the National High-Performance Computing Center at KIT (NHR@KIT). This center is jointly supported by the Federal Ministry of Education and Research and the Ministry of Science, Research and the Arts of Baden-Württemberg, as part of the National High-Performance Computing (NHR) joint funding program (\url{https://www.nhr-verein.de/en/our-partners}). HoreKa is partly funded by the German Research Foundation (DFG). This work was performed with the help of the Large Scale Data Facility at the Karlsruhe Institute of Technology funded by the Ministry of Science, Research and the Arts Baden-Württemberg and by the Federal Ministry of Education and Research. This research was also funded by the Deutsche Forschungsgemeinschaft (DFG, German Research Foundation) via the Project ApplFM (528483508).

{
    \small
    \bibliographystyle{ieeenat_fullname}
    \bibliography{main}
}

\clearpage
\appendix
\input{supp_document}



\end{document}

%% file: content/tables/codebook_eval.tex
\begin{figure*}[t]
    \centering
    \begin{subfigure}[b]{.33\textwidth}
        \includegraphics[height=0.8\textwidth]{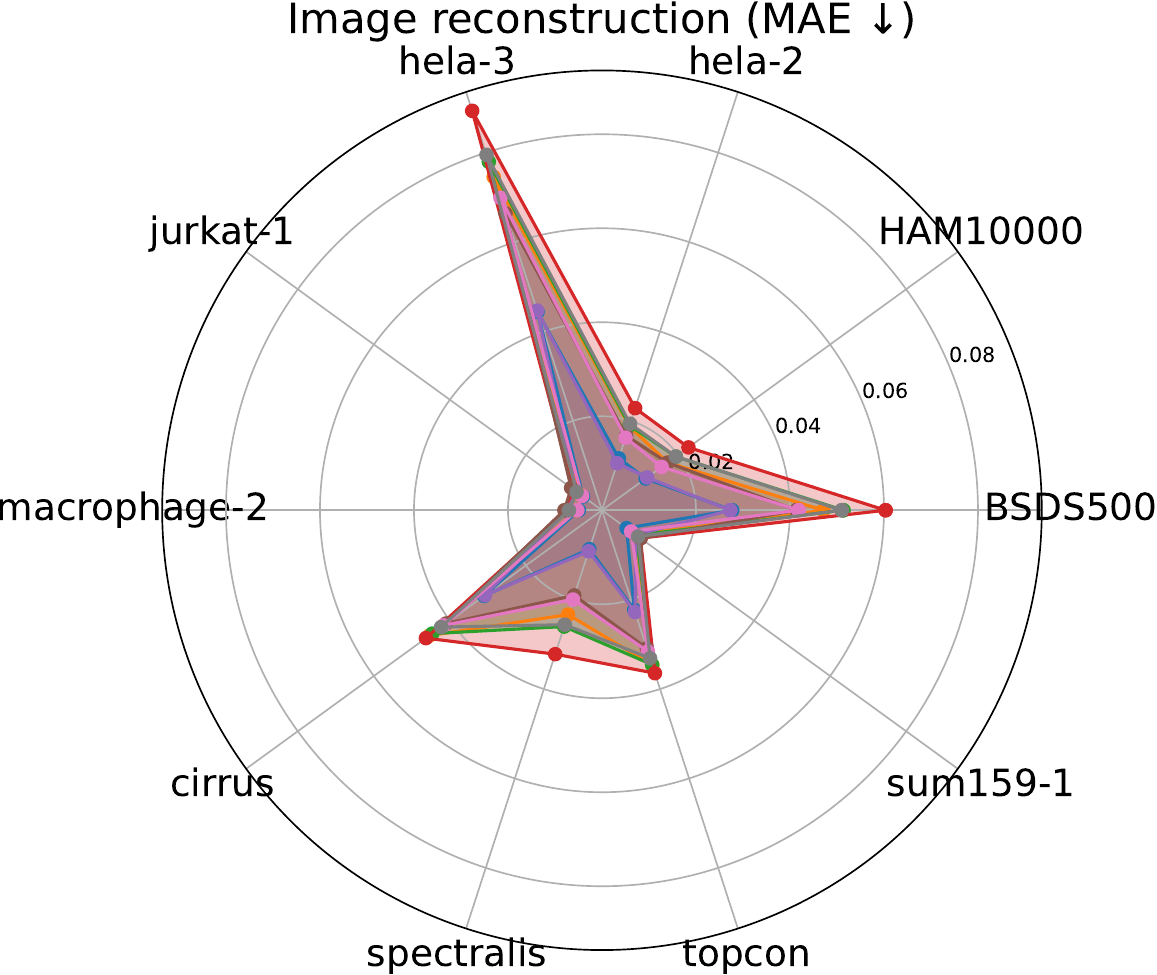}
        \caption{Image reconstruction.}
        \label{fig:spiderplot1}
    \end{subfigure}\begin{subfigure}[b]{.33\textwidth}
        \includegraphics[height=0.77\textwidth]{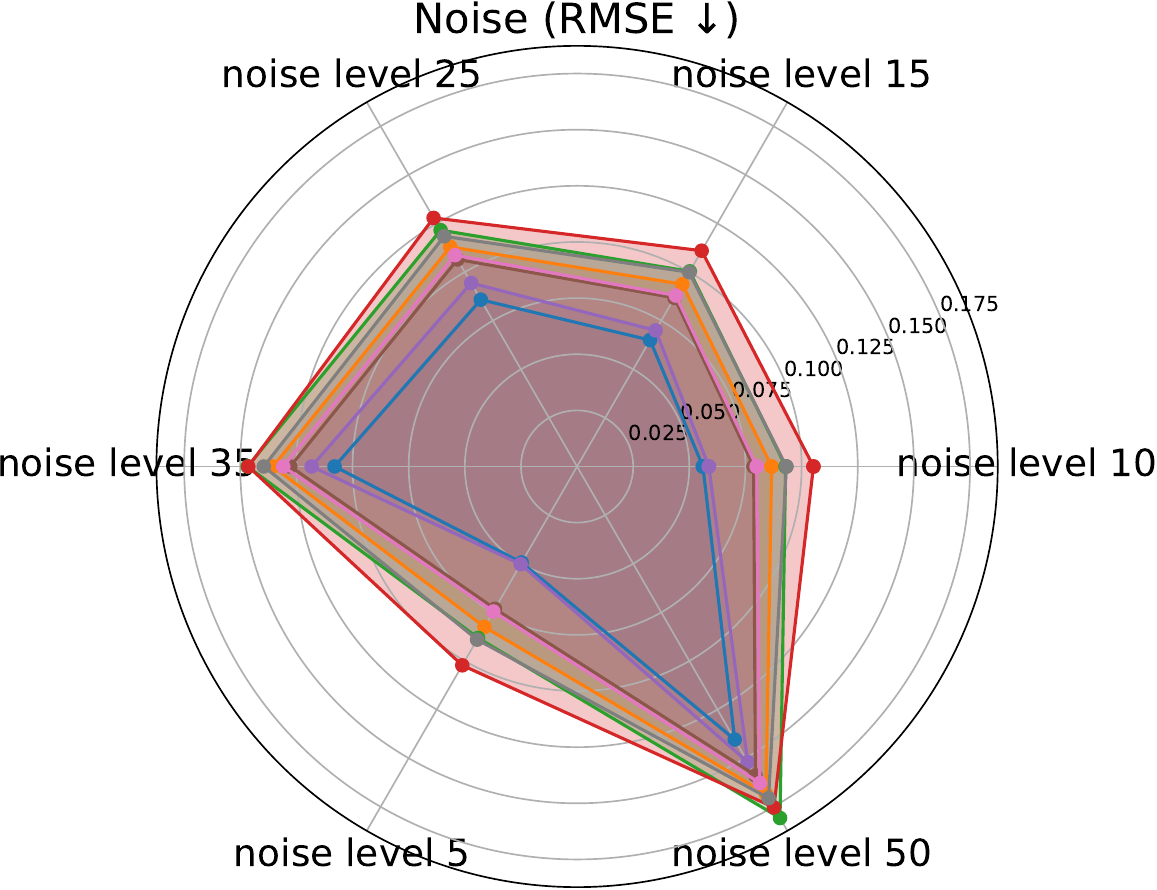}
        \caption{Image compression and denoising.}
        \label{fig:spiderplot2}
    \end{subfigure}\begin{subfigure}[b]{.33\textwidth}
        \includegraphics[height=0.8\textwidth]{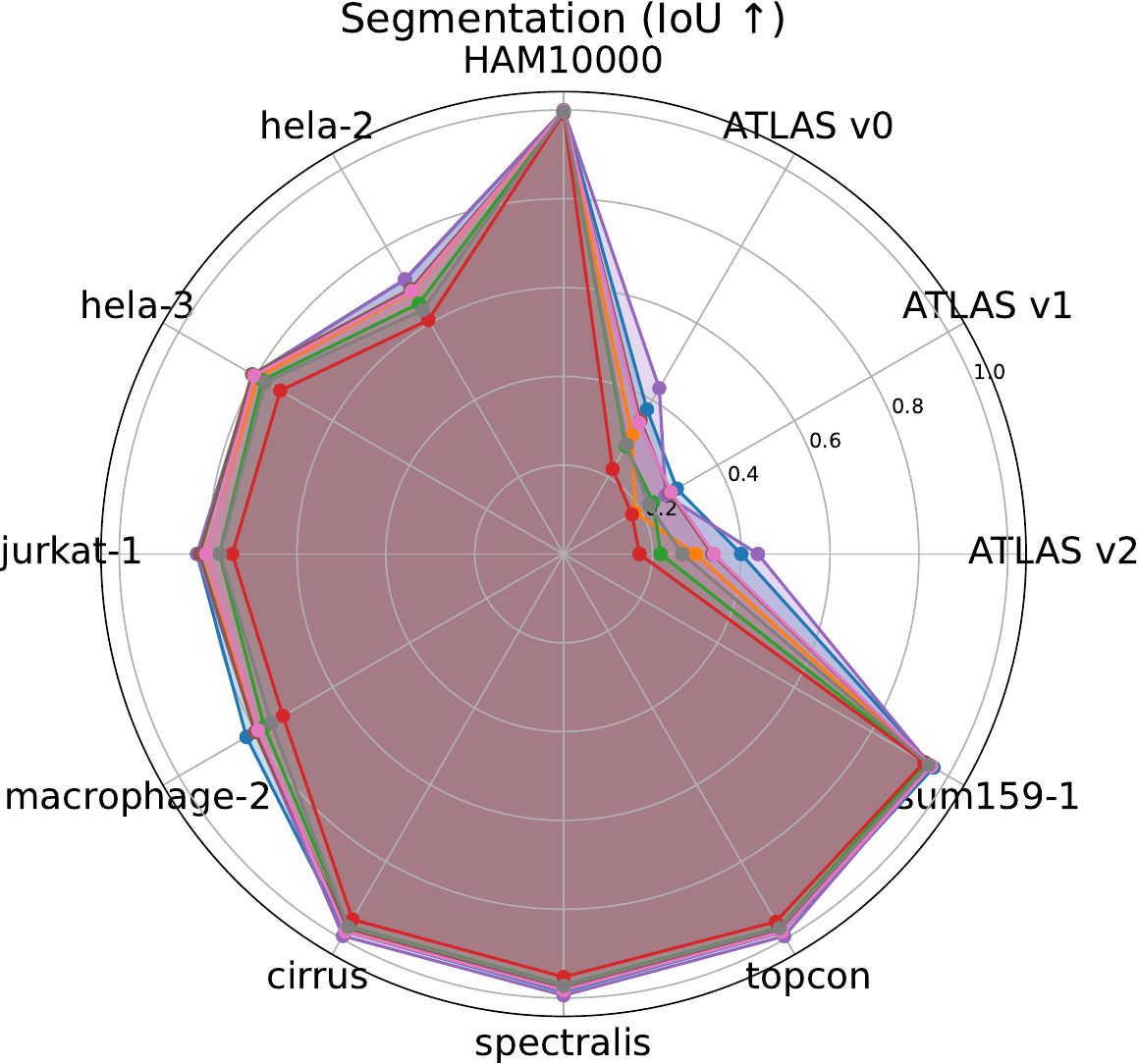}
        \caption{Semantic segmentation.}
        \label{fig:spiderplot0}
    \end{subfigure}
    \includegraphics[width=0.5\linewidth]{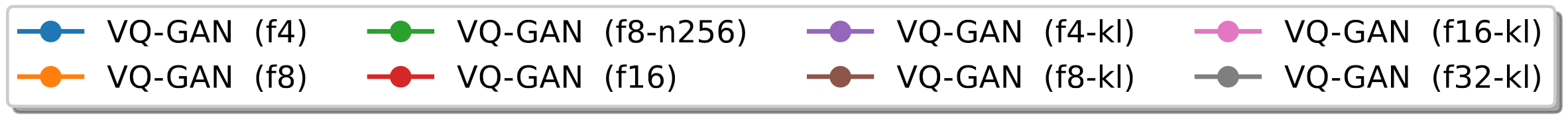}
    \caption{Investigation on how well pre-trained codebooks represent task data. Left to right: Evaluation of image reconstruction, evaluation of reconstruction of different noise levels in BSDS500 and reconstruction efficacy on segmentation maps of different datasets.}
    \label{fig:codebook_spiderplot}
\end{figure*}

%% file: supp_document.tex
\section{Dataset}
\label{sec:dataset}

Our training dataset is based on the MedSAM dataset~\cite{ma2024segment}, more specifically, the $38$ datasets of which we show examples in~\Cref{tab:example_images}, more details can be found in~\Cref{tab:dataset_license}.
As can be seen, the datasets cover a wide variety of imaging modalities.

We enrich the data with synthetic tasks, which we put into task sequences for compositional medical tasks as we described in the main paper.
An example of the individual generative-, transformation- and discriminative tasks we extract from each image-segmentation pair is shown in~\cref{tab:enriched_images_examples}.

\begin{table*}
\small
\centering
\setlength{\tabcolsep}{2pt}
\begin{tabular}{ccccc}
\toprule \\
WMH FLAIR & SpineMR & ProstateADC & COVID-19-Radiogr.-Datab. & Intraretinal-Cystoid-Fluid \\                                                                                          
\includegraphics[width=0.1\textwidth]{ 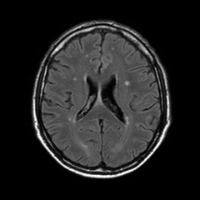}\hspace{0.0004\textwidth}\includegraphics[width=0.1\textwidth]{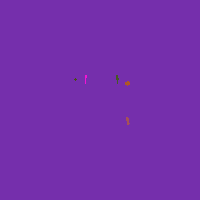}  & \includegraphics[width=0.1\textwidth]{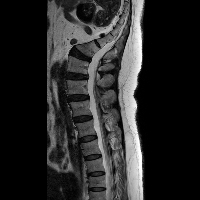}\hspace{0.0004\textwidth}\includegraphics[width=0.1\textwidth]{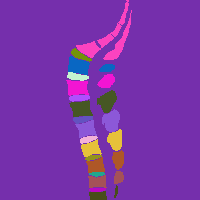}  & \includegraphics[width=0.1\textwidth]{ 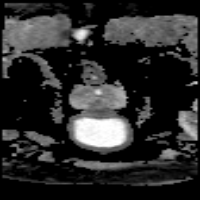 }\hspace{0.0004\textwidth}\includegraphics[width=0.1\textwidth]{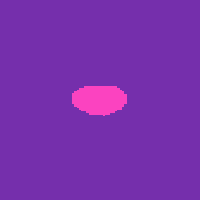}  & \includegraphics[width=0.1\textwidth]{ 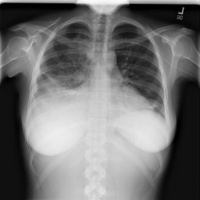 }\hspace{0.0004\textwidth}\includegraphics[width=0.1\textwidth]{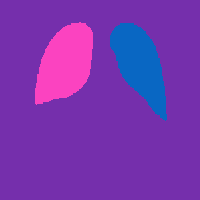 }  & \includegraphics[width=0.1\textwidth]{ 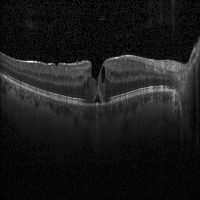 }\hspace{0.0004\textwidth}\includegraphics[width=0.1\textwidth]{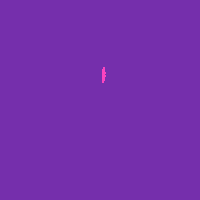 }  \\             
\midrule \\                                                                                                                                                                                
BraTS T1CE & ISLES2022 DWI & COVID19-CT-Seg-Bench & ISIC2018 & m2caiSeg \\

\includegraphics[width=0.1\textwidth]{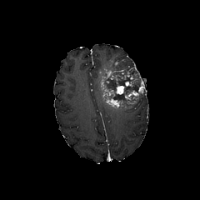}\hspace{0.0004\textwidth}\includegraphics[width=0.1\textwidth]{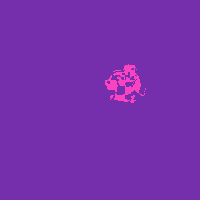 }  & \includegraphics[width=0.1\textwidth]{ 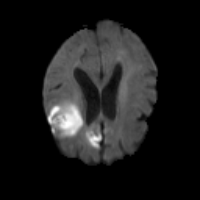}\hspace{0.0004\textwidth}\includegraphics[width=0.1\textwidth]{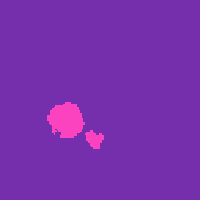}  & \includegraphics[width=0.1\textwidth]{ 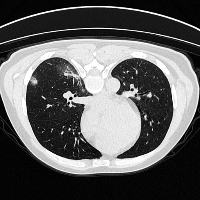}\hspace{0.0004\textwidth}\includegraphics[width=0.1\textwidth]{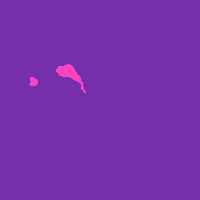 }  & \includegraphics[width=0.1\textwidth]{ 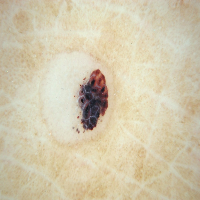 }\hspace{0.0004\textwidth}\includegraphics[width=0.1\textwidth]{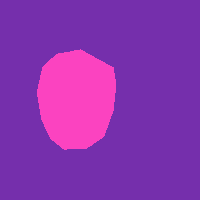 }  & \includegraphics[width=0.1\textwidth]{ 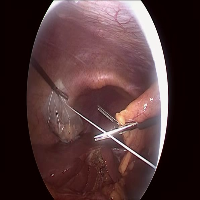 }\hspace{0.0004\textwidth}\includegraphics[width=0.1\textwidth]{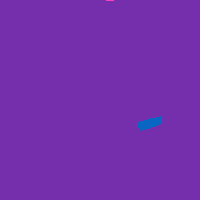 } \\
\midrule \\

AbdomenCT1K & AMOS & ProstateT2 & QIN-PROSTATE-Prostate & LungMasks \\
\includegraphics[width=0.1\textwidth]{ 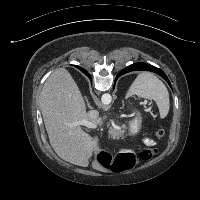 }\hspace{0.0004\textwidth}\includegraphics[width=0.1\textwidth]{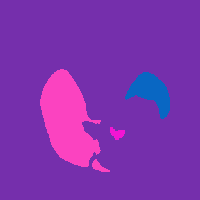 }  & \includegraphics[width=0.1\textwidth]{ 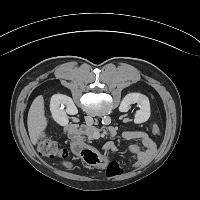 }\hspace{0.0004\textwidth}\includegraphics[width=0.1\textwidth]{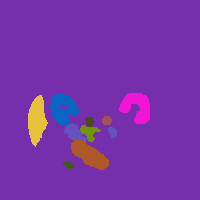 }  & \includegraphics[width=0.1\textwidth]{ 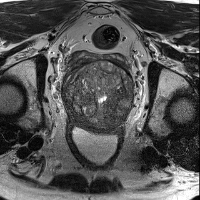 }\hspace{0.0004\textwidth}\includegraphics[width=0.1\textwidth]{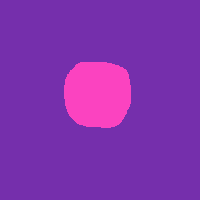 }  & \includegraphics[width=0.1\textwidth]{ 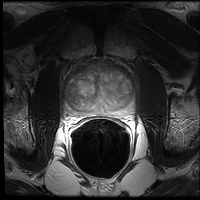 }\hspace{0.0004\textwidth}\includegraphics[width=0.1\textwidth]{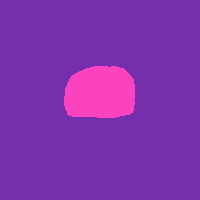 }  & \includegraphics[width=0.1\textwidth]{ 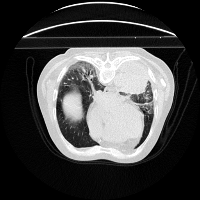 }\hspace{0.0004\textwidth}\includegraphics[width=0.1\textwidth]{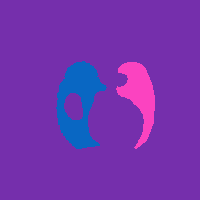 }  \\

\midrule \\

Breast-Ultrasound & Pneumothorax-Masks & Kvasir-SEG & COVID-19-20-CT-SegCh. & LIDC-IDRI \\
\includegraphics[width=0.1\textwidth]{  }\hspace{0.0004\textwidth}\includegraphics[width=0.1\textwidth]{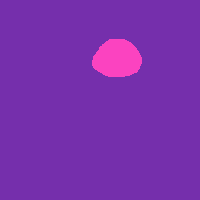 }  & \includegraphics[width=0.1\textwidth]{ 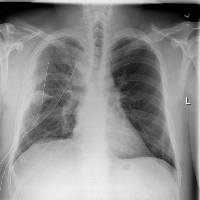 }\hspace{0.0004\textwidth}\includegraphics[width=0.1\textwidth]{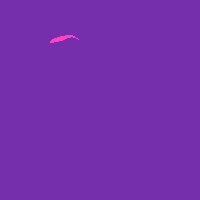 }  & \includegraphics[width=0.1\textwidth]{ 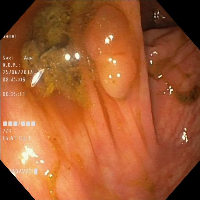 }\hspace{0.0004\textwidth}\includegraphics[width=0.1\textwidth]{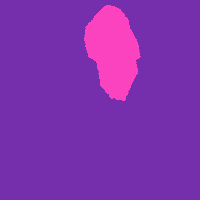 }  & \includegraphics[width=0.1\textwidth]{ 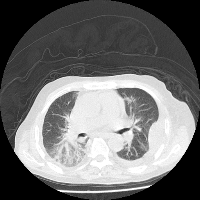 }\hspace{0.0004\textwidth}\includegraphics[width=0.1\textwidth]{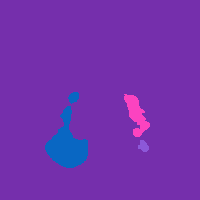 }  & \includegraphics[width=0.1\textwidth]{ 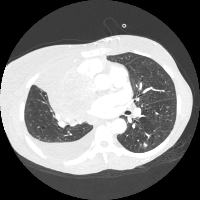 }\hspace{0.0004\textwidth}\includegraphics[width=0.1\textwidth]{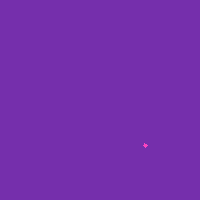 }  \\
\midrule \\
LungLesions & TotalSeg muscles & TotalSeg organs & CT-ORG & Heart \\
\includegraphics[width=0.1\textwidth]{ 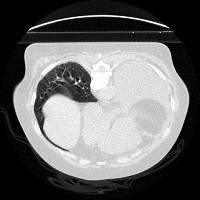 }\hspace{0.0004\textwidth}\includegraphics[width=0.1\textwidth]{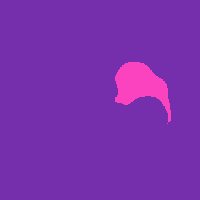 }  & \includegraphics[width=0.1\textwidth]{ 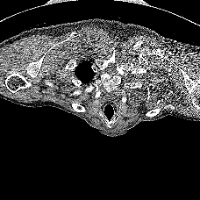 }\hspace{0.0004\textwidth}\includegraphics[width=0.1\textwidth]{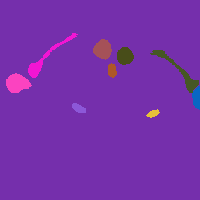 }  & \includegraphics[width=0.1\textwidth]{ 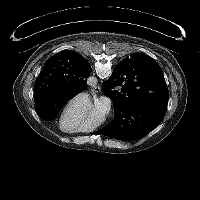 }\hspace{0.0004\textwidth}\includegraphics[width=0.1\textwidth]{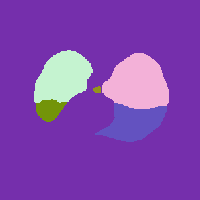 }  & \includegraphics[width=0.1\textwidth]{ 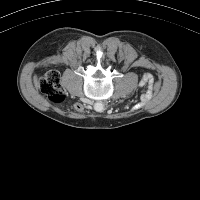 }\hspace{0.0004\textwidth}\includegraphics[width=0.1\textwidth]{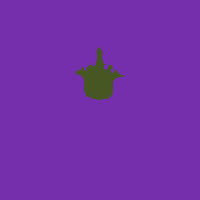 }  & \includegraphics[width=0.1\textwidth]{ 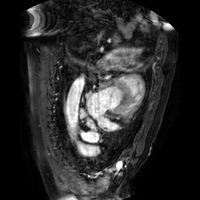 }\hspace{0.0004\textwidth}\includegraphics[width=0.1\textwidth]{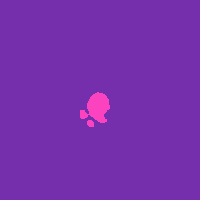 }  \\
\midrule \\
hc18 & TotalSeg cardiac & AMOSMR & BraTS T1 & QIN-PROSTATE-Lesion \\
\includegraphics[width=0.1\textwidth]{ 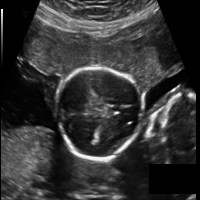 }\hspace{0.0004\textwidth}\includegraphics[width=0.1\textwidth]{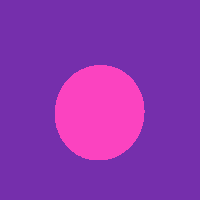 }  & \includegraphics[width=0.1\textwidth]{ 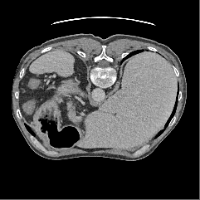 }\hspace{0.0004\textwidth}\includegraphics[width=0.1\textwidth]{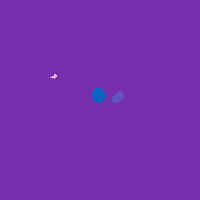 }  & \includegraphics[width=0.1\textwidth]{ 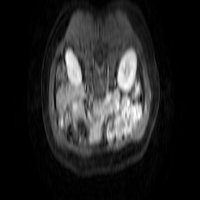 }\hspace{0.0004\textwidth}\includegraphics[width=0.1\textwidth]{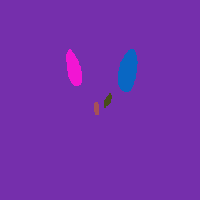 }  & \includegraphics[width=0.1\textwidth]{ 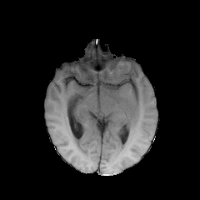 }\hspace{0.0004\textwidth}\includegraphics[width=0.1\textwidth]{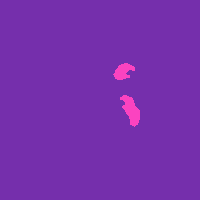 }  & \includegraphics[width=0.1\textwidth]{ 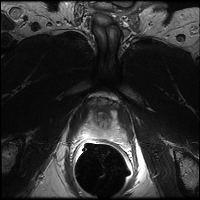 }\hspace{0.0004\textwidth}\includegraphics[width=0.1\textwidth]{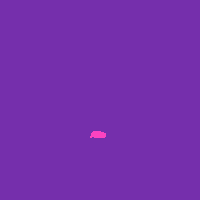 }  \\

\midrule \\

Chest-Xray-Masks-Labels & CDD-CESM & TCIA-LCTSC & ISLES2022 ADC & COVID-QU-Ex-l.M. Lung \\

\includegraphics[width=0.1\textwidth]{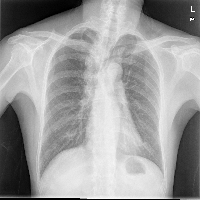}\hspace{0.0004\textwidth}\includegraphics[width=0.1\textwidth]{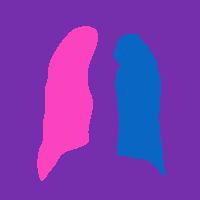}  & \includegraphics[width=0.1\textwidth]{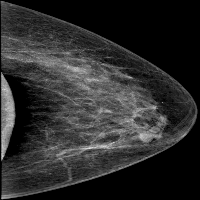}\hspace{0.0004\textwidth}\includegraphics[width=0.1\textwidth]{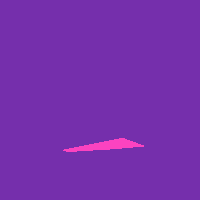}  & \includegraphics[width=0.1\textwidth]{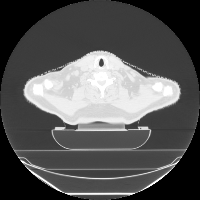}\hspace{0.0004\textwidth}\includegraphics[width=0.1\textwidth]{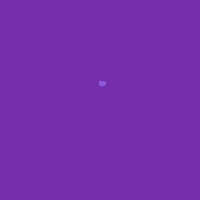}  & \includegraphics[width=0.1\textwidth]{ 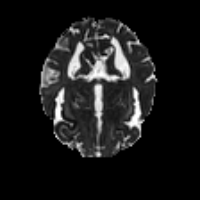 }\hspace{0.0004\textwidth}\includegraphics[width=0.1\textwidth]{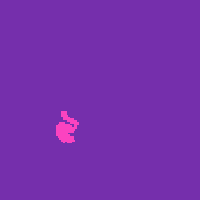 }  & \includegraphics[width=0.1\textwidth]{ 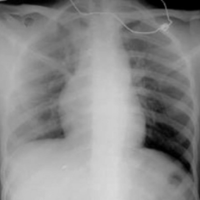 }\hspace{0.0004\textwidth}\includegraphics[width=0.1\textwidth]{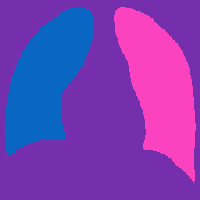 }  \\

CholecSeg8k & crossmoda & BraTS FLAIR & & \\
\includegraphics[width=0.1\textwidth]{ 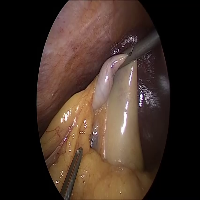 }\hspace{0.0004\textwidth}\includegraphics[width=0.1\textwidth]{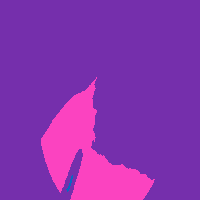 }  & \includegraphics[width=0.1\textwidth]{ 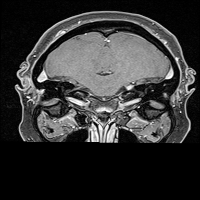 }\hspace{0.0004\textwidth}\includegraphics[width=0.1\textwidth]{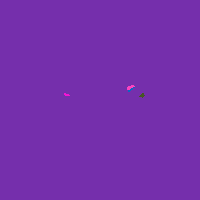 }  & \includegraphics[width=0.1\textwidth]{ 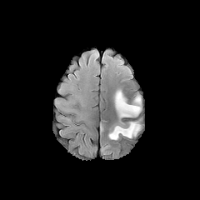 }\hspace{0.0004\textwidth}\includegraphics[width=0.1\textwidth]{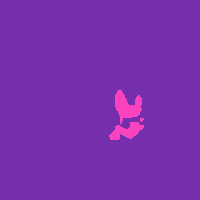 } & & \\
\bottomrule
\end{tabular}
\caption{Example images and segmentations from $38$ datasets in MedSAM~\cite{ma2024segment} which we use as basis for our investigation.}

\label{tab:example_images}
\end{table*}

{
\begin{table*}
\scriptsize
\begin{tabular}{ll}
    \toprule
    Dataset & Link\\
    \midrule
    AMOSMR~\cite{ji2022amos} & \url{https://amos22.grand-challenge.org/Dataset/}\\
    BraTS T1~\cite{bakas2017segmentation,bakas10segmentation,bakas2017advancing,bakas2018identifying,menze2014multimodal} & \url{http://braintumorsegmentation.org/}\\
    BraTS T1CE~\cite{bakas2017segmentation,bakas10segmentation,bakas2017advancing,bakas2018identifying,menze2014multimodal} & \url{http://braintumorsegmentation.org/} \\
    CDD-CESM~\cite{khaled2021categorized} & \url{https://www.cancerimagingarchive.net/collection/cdd-cesm/} \\
    Chest-Xray-Masks-Labels~\cite{jaeger2013automatic,candemir2013lung}  & \url{https://www.kaggle.com/datasets/nikhilpandey360/chest-xray-masks-and-labels} \\
    COVID-19-20-CT-SegCh.~\cite{roth2022rapid,ct_images}  & \url{https://covid-segmentation.grand-challenge.org/Data/} \\
    COVID-QU-Ex-l.M. Lung.~\cite{tahir2021covid,rahman2021exploring,degerli2021covid,chowdhury2020can,anas_m__tahir_muhammad_e__h__chowdhury_yazan_qiblawey_amith_khandakar_tawsifur_rahman_serkan_kiranyaz_uzair_khurshid_nabil_ibtehaz_sakib_mahmud_maymouna_ezeddin_2022}  & \url{https://www.kaggle.com/datasets/anasmohammedtahir/covidqu} \\
    COVID19-CT-Seg-Bench~\cite{ma2021toward}  & \url{https://github.com/JunMa11/COVID-19-CT-Seg-Benchmark} \\
    crossmoda~\cite{dorent2023crossmoda} & \url{https://crossmoda-challenge.ml/} \\
    hc18~\cite{van2018automated}  & \url{https://hc18.grand-challenge.org/} \\
    Heart~\cite{simpson2019large}  & \url{http://medicaldecathlon.com/} \\
    Intraretinal-Cystoid-Fluid~\cite{ahmed2022deep}  & \url{https://www.kaggle.com/datasets/zeeshanahmed13/intraretinal-cystoid-fluid} \\
    ISLES2022 ADC~\cite{hernandez2022isles}  & \url{https://www.isles-challenge.org/} \\
    ISLES2022 DWI~\cite{hernandez2022isles}  & \url{https://www.isles-challenge.org/} \\
    Kvasir-SEG~\cite{jha2019kvasir,pogorelov2017kvasir}  & \url{https://datasets.simula.no/kvasir/} \\
    LungLesions~\cite{antonelli2022medical}  & \url{http://medicaldecathlon.com/} \\
    m2caiSeg~\cite{maqbool2020m2caiseg}  & \url{https://www.kaggle.com/datasets/salmanmaq/m2caiseg} \\
    Pneumothorax-Masks~\cite{siim-acr-pneumothorax-segmentation}  & \url{https://www.kaggle.com/datasets/vbookshelf/pneumothorax-chest-xray-images-and-masks} \\
    ProstateADC~\cite{simpson2019large}  & \url{http://medicaldecathlon.com} \\
    ProstateT2~\cite{simpson2019large}  & \url{http://medicaldecathlon.com} \\
    QIN-PROSTATE-Lesion~\cite{antonelli2022medical,fedorov2018annotated} & \url{http://doi.org/10.7937/K9/TCIA.2018.MR1CKGND} \\
    QIN-PROSTATE-Prostate~\cite{antonelli2022medical,fedorov2018annotated}  & \url{http://doi.org/10.7937/K9/TCIA.2018.MR1CKGND} \\
    SpineMR~\cite{zukic2014robust}  & \url{https://www.cg.informatik.uni-siegen.de/en/spine-segmentation-and-analysis} \\
    TCIA-LCTSC~\cite{lctsc}  & \url{https://www.cancerimagingarchive.net/collection/lctsc/} \\
    WMH flair~\cite{kuijf2019standardized} & \url{https://wmh.isi.uu.nl/} \\
    BraTS flair~\cite{bakas2017segmentation,bakas10segmentation,bakas2017advancing,bakas2018identifying,menze2014multimodal} & \url{http://braintumorsegmentation.org/} \\
    Breast-Ultrasound~\cite{al2020dataset} & \url{https://www.kaggle.com/datasets/aryashah2k/breast-ultrasound-images-dataset} \\
    COVID-19-Radiogr.-Datab.~\cite{rahman2021exploring,chowdhury2020can}  & \url{https://www.kaggle.com/datasets/tawsifurrahman/covid19-radiography-database} \\
    ISIC2018~\cite{tschandl2018ham10000,codella2018skin,codella2019skin} & \url{https://challenge.isic-archive.com/data/} \\
    CholecSeg8k~\cite{hong2020cholecseg8k,twinanda2016endonet} & \url{https://www.kaggle.com/datasets/newslab/cholecseg8k} \\
    TotalSeg cardiac~\cite{wasserthal2023totalsegmentator}  & \url{https://zenodo.org/records/6802614} \\
    CT-ORG~\cite{rister2020ct} & \url{https://www.cancerimagingarchive.net/collection/ct-org/} \\
    TotalSeg organs~\cite{wasserthal2023totalsegmentator} & \url{https://zenodo.org/records/6802614} \\
    TotalSeg muscles~\cite{wasserthal2023totalsegmentator} & \url{https://zenodo.org/records/6802614} \\
    LIDC-IDRI~\cite{armato2015data} & \url{https://www.cancerimagingarchive.net/collection/lidc-idri/} \\
    AMOS~\cite{ji2022amos} & \url{https://amos22.grand-challenge.org/} \\
    AbdomenCT1K~\cite{Ma-2021-AbdomenCT-1K,ma2022fast} & \url{https://github.com/JunMa11/AbdomenCT-1K} \\
    LungMasks~\cite{hofmanninger2020automatic} & \url{https://github.com/JoHof/lungmask} \\
    \bottomrule
\end{tabular}
\caption{Datasets with their citations and links for further information.}
\label{tab:dataset_license}
\end{table*}
}

\section{Information on experiments in \textit{Chapter 4. Analysis of task recovery in codebooks}}

{
\begin{table*}
\tiny
\centering
\setlength{\tabcolsep}{1pt}
\begin{tabular}{cccccccccc}
\toprule \\
box (width 1) & box (width 1) & box (width 1) & box (width 3) & box (width 3)  & box (width 3)  & box (width 5)  & box (width 5) & box (width 5) & brightness \\
\includegraphics[width=0.09\textwidth]{ 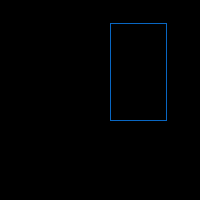 }  & \includegraphics[width=0.09\textwidth]{ 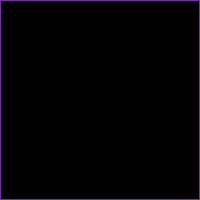 }  & \includegraphics[width=0.09\textwidth]{ 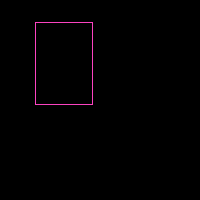 }  & \includegraphics[width=0.09\textwidth]{ 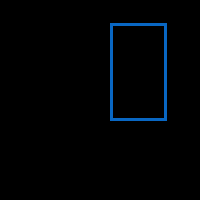 }  & \includegraphics[width=0.09\textwidth]{ 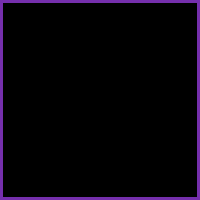 }  & \includegraphics[width=0.09\textwidth]{ 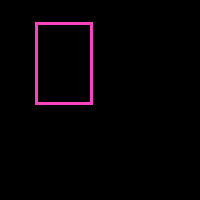 }  & \includegraphics[width=0.09\textwidth]{ 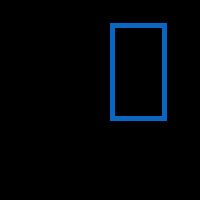 }  & \includegraphics[width=0.09\textwidth]{ 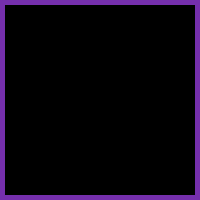 }  & \includegraphics[width=0.09\textwidth]{ 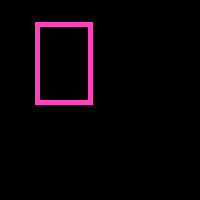 }  & \includegraphics[width=0.09\textwidth]{ 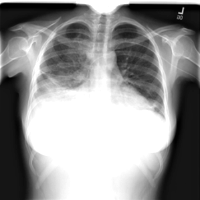 }  \\
\midrule \\
color jitter & noise & equalize & h-flip & image & inv-intens. & inpaint & point (width 1) & point (width 1) & point (width 1) \\
\includegraphics[width=0.09\textwidth]{ 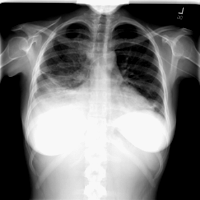 }  & \includegraphics[width=0.09\textwidth]{ 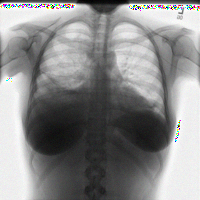 }  & \includegraphics[width=0.09\textwidth]{ 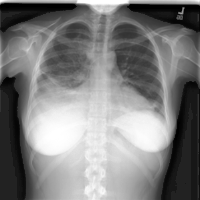 }  & \includegraphics[width=0.09\textwidth]{ 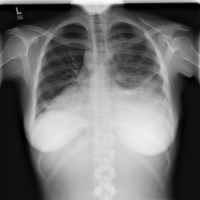 }  & \includegraphics[width=0.09\textwidth]{ 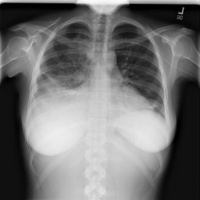 }  & \includegraphics[width=0.09\textwidth]{ 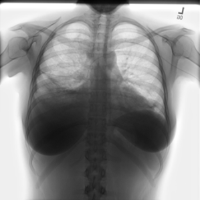 }  & \includegraphics[width=0.09\textwidth]{ 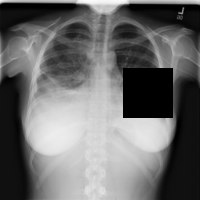 }  & \includegraphics[width=0.09\textwidth]{ 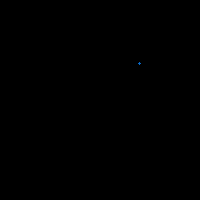 }  & \includegraphics[width=0.09\textwidth]{ 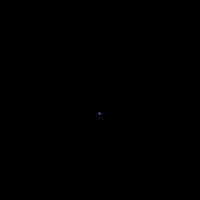 }  & \includegraphics[width=0.09\textwidth]{ 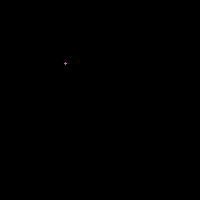 }  \\
\midrule \\
point (width 3)&point (width 3)& point (width 3) &point (width 5) & point (width 5) & point (width 5) & rotate $180^\circ$ &  rotate $270^\circ$ &  rotate $90^\circ$ & segmentation \\
\includegraphics[width=0.09\textwidth]{ 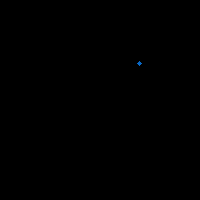 }  & \includegraphics[width=0.09\textwidth]{ 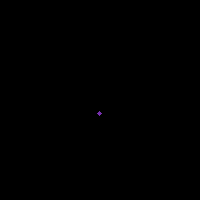 }  & \includegraphics[width=0.09\textwidth]{ 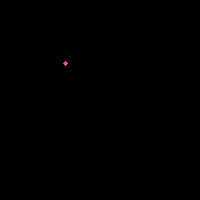 }  & \includegraphics[width=0.09\textwidth]{ 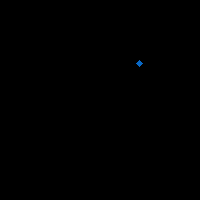 }  & \includegraphics[width=0.09\textwidth]{ 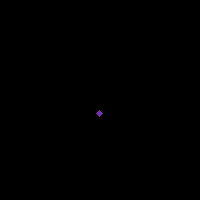 }  & \includegraphics[width=0.09\textwidth]{ 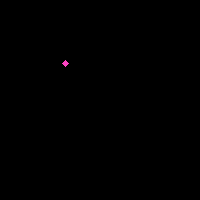 }  & \includegraphics[width=0.09\textwidth]{ 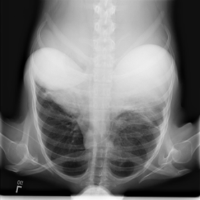 }  & \includegraphics[width=0.09\textwidth]{ 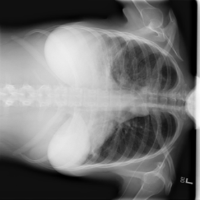 }  & \includegraphics[width=0.09\textwidth]{ 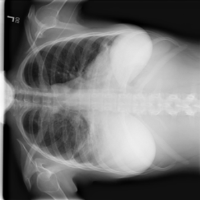 }  & \includegraphics[width=0.09\textwidth]{ 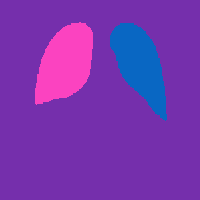 }  \\
\midrule \\
edges (width 1) & edges (width 1) & edges (width 1) & edges (width 3) &edges (width 3) &edges (width 3) & edges (width 5) & edges (width 5) &  edges (width 5) & skeleton (width 1) \\
\includegraphics[width=0.09\textwidth]{ 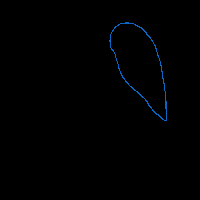 }  & \includegraphics[width=0.09\textwidth]{ 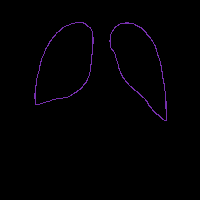 }  & \includegraphics[width=0.09\textwidth]{ 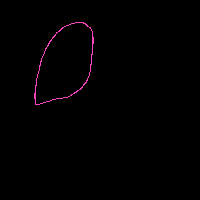 }  & \includegraphics[width=0.09\textwidth]{ 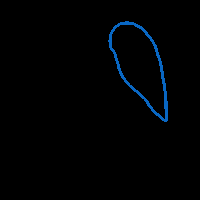 }  & \includegraphics[width=0.09\textwidth]{ 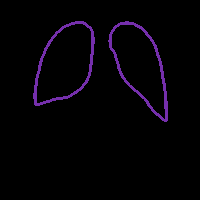 }  & \includegraphics[width=0.09\textwidth]{ 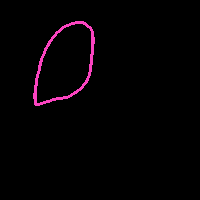 }  & \includegraphics[width=0.09\textwidth]{ 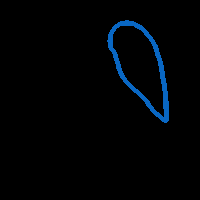 }  & \includegraphics[width=0.09\textwidth]{ 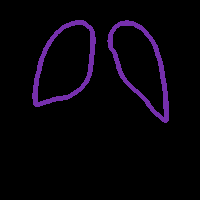 }  & \includegraphics[width=0.09\textwidth]{ 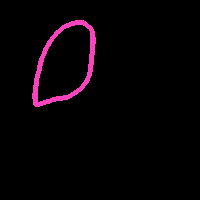 }  & \includegraphics[width=0.09\textwidth]{ 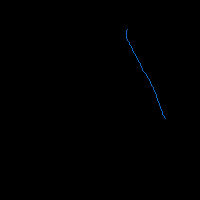 }  \\
\midrule \\
skeleton (width 1) & skeleton (width 1) & skeleton (width 3) & skeleton (width 3) & skeleton (width 3) & skeleton (width 5) & skeleton (width 5) & skeleton (width 5) & super res. 100 & super res. 25 \\
\includegraphics[width=0.09\textwidth]{ 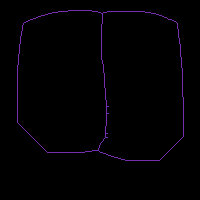 }  & \includegraphics[width=0.09\textwidth]{ 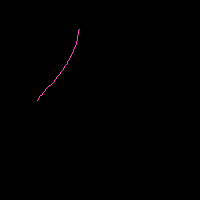 }  & \includegraphics[width=0.09\textwidth]{ 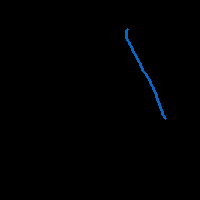 }  & \includegraphics[width=0.09\textwidth]{ 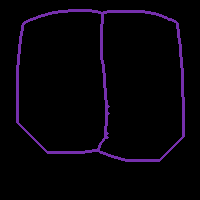 }  & \includegraphics[width=0.09\textwidth]{ 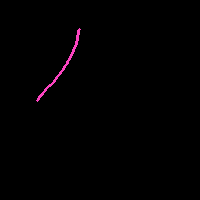 }  & \includegraphics[width=0.09\textwidth]{ 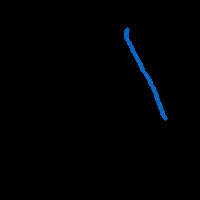 }  & \includegraphics[width=0.09\textwidth]{ 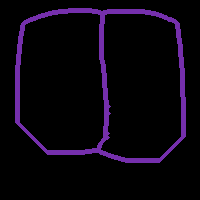 }  & \includegraphics[width=0.09\textwidth]{ 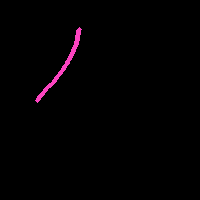 }  & \includegraphics[width=0.09\textwidth]{ 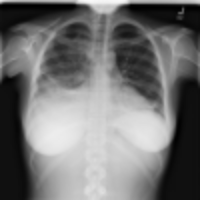 }  & \includegraphics[width=0.09\textwidth]{ 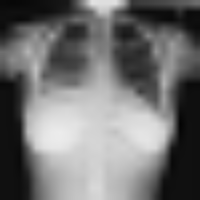 }  \\
\midrule \\
super res. 50 & v-flip \\
\includegraphics[width=0.09\textwidth]{ 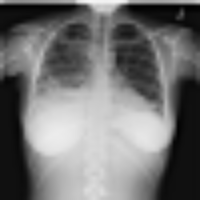 }  & \includegraphics[width=0.09\textwidth]{ 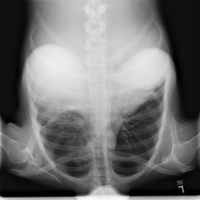 }  \\

\bottomrule
\end{tabular}
\caption{Here, we show an example of how each image-segmentation pair is enriched with additional synthetic tasks.}
\label{tab:enriched_images_examples}
\end{table*}
}

As we oultine in the main paper, we identify the limitations that are induced when operating in the token space of learned codebooks.
These codebooks entail certain reconstruction errors, which we explored for simple image reconstruction and the reconstruction of segmentation maps on the datasets DAP ATLAS~\cite{jaus2023towards}, BSDS/CBSD68~\cite{MartinFTM01}, HAM10000~\cite{tschandl2018ham10000}, Openorganelle (hela-2, hela-3, jurkat-1, macrophage-2, sum-159)~\cite{heinrich2021whole} and RETOUCH (cirrus, topcon, spectralis)~\cite{bogunovic2019retouch}.
In~\Cref{tab:codebook_eval2}, we show the results from the spiderplots in the main paper in numerical form.
There, results for auto-encoding images, noisy images and segmentation maps are reported.
For the ATLAS dataset with its $142$ classes, we briefly investigated the significance of different color schemes when processing segmentation maps.
To do this, we utilized color-maps from matplotlib~\footnote{\url{https://matplotlib.org/stable/users/explain/colors/colormaps.html}}, the names of which are indicated in~\Cref{tab:codebook_eval2}.
In~\Cref{fig:visualprompts} we show a segmentation map with these different color-maps.

{
\begin{figure}
    \centering
    \setlength{\tabcolsep}{.7pt}

    \begin{tabular}{ccc}
        \includegraphics[width=0.33\linewidth]{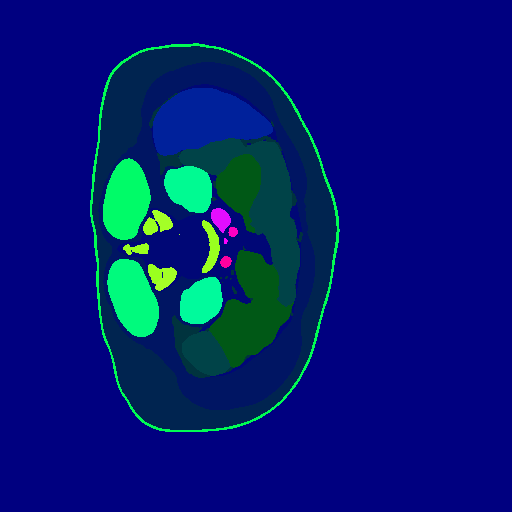}&    \includegraphics[width=0.33\linewidth]{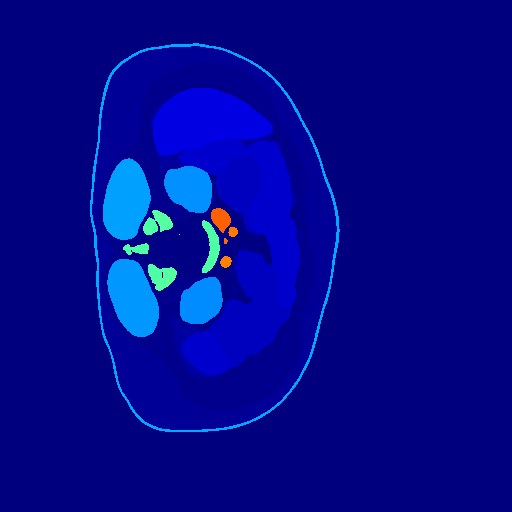}&    \includegraphics[width=0.33\linewidth]{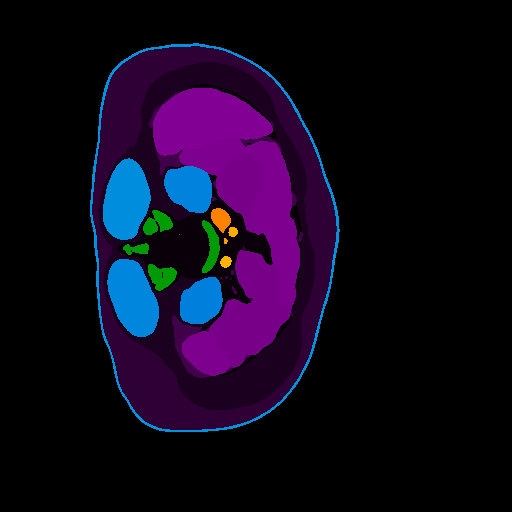} \\
         gist ncar & jet & nipy spectral
    \end{tabular}
    \caption{Different colormaps for the ATLAS dataset with which we supply codebooks to indicate the effect of visual prompting.}
    \label{fig:visualprompts}
\end{figure}
}

The different VQ-GAN models we test are pre-trained models from a model zoo~\footnote{\url{https://github.com/CompVis/latent-diffusion/tree/main?tab=readme-ov-file\#pretrained-autoencoding-models}} which is associated to~\cite{rombach2021highresolution,Retrieval}.
The models vary in their loss function in training and network architectures which influences their codebook size and latent tensor shape in the quantization layer.
For a detailed description of their differences we kindly refer the reader to the explanations by Rombach \etal~\cite{rombach2021highresolution}.

\section{Hyperparameters for VQ-GAN and tranformer training}
In~\Cref{tab:hyperparams}, we summarize the genreal hyperparameters of the VQ-GAN models and the transformer models we train.
Both architectures and the effects of the hyperparameters in source code can be found in the model definitions of either VQ-GAN~\footnote{\url{https://github.com/CompVis/taming-transformers/blob/master/taming/models/vqgan.py}} or the GPT2 transformer~\footnote{\url{https://github.com/CompVis/taming-transformers/blob/master/taming/modules/transformer/mingpt.py}}.

\section{Additional qualitative results}

\subsection{VQ-GAN reconstruction}

\begin{figure}
    \centering
    \rotatebox[origin=l]{90}{inputs}\includegraphics[width=0.95\linewidth]{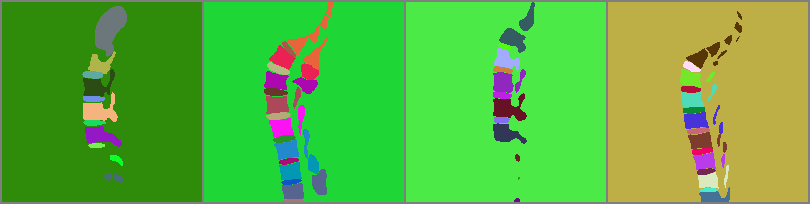}
    
    \rotatebox[origin=l]{90}{pre-trained}\includegraphics[width=0.95\linewidth]{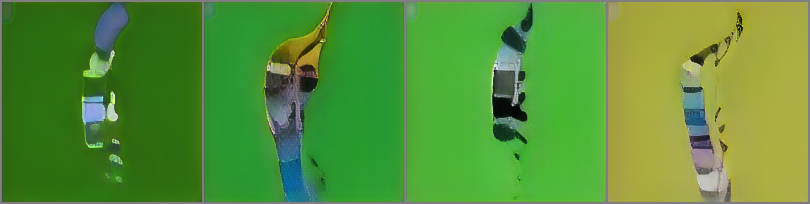}
    
    \rotatebox[origin=l]{90}{fine-tuned} \includegraphics[width=0.95\linewidth]{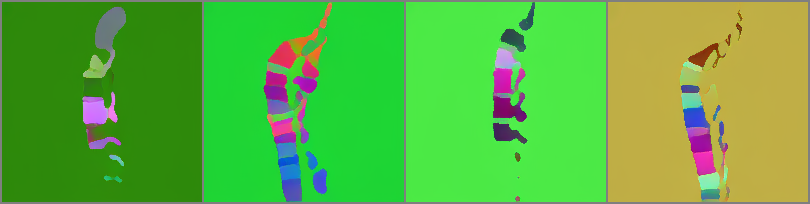}
    \caption{Qualitative examples of auto-encoding segmentation maps with VQ-GAN codebooks. First row shows inputs, second row shows reconstructions with an ImageNet pre-trained VQ-GAN, row three shows a model fine-tuned on task data with color remapping augmentation and dataset + task balancing.}
    \label{fig:vqganqualitative}
\end{figure}

In the main paper, we present quantitative results regarding the upper bound of codebooks on the MedSAM datasets.
There, we evaluate a pre-trained VQ-GAN and different fine-tuned variants which are trained on task data and in-domain images.
In~\Cref{fig:vqganqualitative}, we show the difference, that our proposed fine-tuning (color remapping augmentation, dataset- and task balancing) can make for representing task data as compared to an ImageNet pre-trained model.
It is clearly evident, that the pre-trained reconstructions are not able to capture the semantic content encoded in the segmentation maps, while fine-tuning shows visually coherent results with only minor inaccuracies in color and shape.

The quantitative evaluation for reconstructing different task-related images (\ie, Table 2 in the main paper) is carried out on $22$K test images and segmentation annotations comprised of samples from the $38$ datasets we use to train.

{
\begin{table*}[t]
    \centering
    \tiny
    \setlength{\tabcolsep}{.7pt}
    \begin{tabular}{l|ccccccccccccccccccccccccccccc}
         \toprule
         Codebook & \rotatebox[origin=l]{90}{ATLAS gist ncar (seg)} & \rotatebox[origin=l]{90}{ATLAS jet (seg)} & \rotatebox[origin=l]{90}{ATLAS nipy spectral (seg)} & \rotatebox[origin=l]{90}{BSDS500 (img)}& \rotatebox[origin=l]{90}{CBSD68 (img)} & \rotatebox[origin=l]{90}{CBSD68 (img noisy5)} & \rotatebox[origin=l]{90}{CBSD68 (img noisy10)} & \rotatebox[origin=l]{90}{CBSD68 (img noisy15)} & \rotatebox[origin=l]{90}{CBSD68 (img noisy25)} & \rotatebox[origin=l]{90}{CBSD68 (img noisy35)} & \rotatebox[origin=l]{90}{CBSD68 (img noisy50)} & \rotatebox[origin=l]{90}{HAM10000 (img)} & \rotatebox[origin=l]{90}{HAM10000 (seg)} & \rotatebox[origin=l]{90}{hela-2 (img)} & \rotatebox[origin=l]{90}{hela-2 (seg)} & \rotatebox[origin=l]{90}{hela-3 (img)} & \rotatebox[origin=l]{90}{hela-3 (seg)} & \rotatebox[origin=l]{90}{jurkat-1 (img)} & \rotatebox[origin=l]{90}{jurkat-1 (seg)} & \rotatebox[origin=l]{90}{macrophage-2 (img)} & \rotatebox[origin=l]{90}{macrophage-2 (seg)} & \rotatebox[origin=l]{90}{sum159-1 (img)} & \rotatebox[origin=l]{90}{sum159-1 (seg)} & \rotatebox[origin=l]{90}{retouch cirrus (img)} & \rotatebox[origin=l]{90}{retouch cirrus (seg)} & \rotatebox[origin=l]{90}{retouch spectralis (img)} & \rotatebox[origin=l]{90}{retouch spectralis (seg)} & \rotatebox[origin=l]{90}{retouch topcon (img)} & \rotatebox[origin=l]{90}{retouch topcon (seg)} \\
         \midrule
         &IoU$\uparrow$&IoU$\uparrow$&IoU$\uparrow$&MAE$\downarrow$&PSNR$\uparrow$&RMSE$\downarrow$&RMSE$\downarrow$&RMSE$\downarrow$&RMSE$\downarrow$&RMSE$\downarrow$&RMSE$\downarrow$&MAE$\downarrow$&IoU$\uparrow$&MAE$\downarrow$&IoU$\uparrow$&MAE$\downarrow$&IoU$\uparrow$&MAE$\downarrow$&IoU$\uparrow$&MAE$\downarrow$&IoU$\uparrow$&MAE$\downarrow$&IoU$\uparrow$&MAE$\downarrow$&IoU$\uparrow$&MAE$\downarrow$&IoU$\uparrow$&MAE$\downarrow$&IoU$\uparrow$\\
         \midrule
         VQ-GAN f4 &  $0.40$  &  $\textbf{ 0.29 }$  &  $0.38$  &  $0.03$ &  $27.39$  &  $\textbf{ 0.05 }$  &  $\textbf{ 0.06 }$  &  $\textbf{ 0.07 }$  &  $\textbf{ 0.09 }$  &  $\textbf{ 0.11 }$  &  $\textbf{ 0.14 }$  &  $\textbf{ 0.01 }$  &  $1.00$  &  $0.01$  &  $0.71$  &  $\textbf{ 0.04 }$  &  $0.81$  &  $\textbf{ 0.00 }$  &  $\textbf{ 0.83 }$  &  $\textbf{ 0.00 }$  &  $\textbf{ 0.82 }$ &  $\textbf{ 0.01 }$ &  $\textbf{ 0.96 }$ &  $0.03$ &  $0.98$ &  $\textbf{ 0.01 }$ &  $0.99$ &  $\textbf{ 0.02 }$ &  $0.98$ \\
         VQ-GAN f8 &  $0.30$ &  $0.19$ &  $0.31$ &  $0.05$ &  $22.48$ &  $0.08$ &  $0.09$ &  $0.09$ &  $0.11$ &  $0.14$ &  $0.17$ &  $0.02$ &  $1.00$ &  $0.02$ &  $0.68$ &  $0.07$ &  $0.80$ &  $0.01$ &  $0.81$ &  $0.01$ &  $0.80$ &  $0.01$ &  $0.95$ &  $0.05$ &  $0.98$ &  $0.02$ &  $0.98$ &  $0.03$ &  $0.98$ \\
         VQ-GAN f8-n256 &  $0.22$ &  $0.23$ &  $0.28$ &  $0.05$ &  $21.81$ &  $0.09$ &  $0.09$ &  $0.10$ &  $0.12$ &  $0.15$ &  $0.18$ &  $0.02$ &  $1.00$ &  $0.02$ &  $0.65$ &  $0.08$ &  $0.78$ &  $0.01$ &  $0.77$ &  $0.01$ &  $0.78$ &  $0.01$ &  $0.94$ &  $0.04$ &  $0.97$ &  $0.03$ &  $0.97$ &  $0.03$ &  $0.97$ \\
         VQ-GAN f16 &  $0.17$ &  $0.18$ &  $0.22$ &  $0.06$ & $20.50$ &  $0.10$ &  $0.11$ &  $0.11$ &  $0.13$ &  $0.15$ &  $0.18$ &  $0.02$ &  $0.99$ &  $0.02$ &  $0.61$ &  $0.09$ &  $0.74$ &  $0.01$ &  $0.75$ &  $0.01$ &  $0.73$ &  $0.01$ &  $0.94$ &  $0.05$ &  $0.95$ &  $0.03$ &  $0.95$ &  $0.04$ &  $0.96$ \\
         VQ-GAN f4-kl &  $\textbf{ 0.44 }$ &  $0.27$ &  $\textbf{ 0.43 }$ &  $\textbf{ 0.03 }$ &  $\textbf{ 27.46 }$ &  $0.05$ &  $0.06$ &  $0.07$ &  $0.09$ &  $0.12$ &  $0.15$ &  $0.01$ &  $\textbf{ 1.00 }$ &  $\textbf{ 0.01 }$ &  $\textbf{ 0.71 }$ &  $0.04$ &  $0.81$ &  $0.01$ &  $0.82$ &  $0.01$ &  $0.80$ &  $0.01$ &  $0.96$ &  $\textbf{ 0.03 }$ &  $\textbf{ 0.99 }$ &  $0.01$ &  $\textbf{ 0.99 }$ &  $0.02$ &  $\textbf{ 0.99 }$ \\
         VQ-GAN f8-kl &  $0.33$ &  $0.28$ &  $0.35$ &  $0.04$ &  $23.65$ &  $0.07$ &  $0.08$ &  $0.09$ &  $0.11$ &  $0.13$ &  $0.16$ &  $0.02$ &  $1.00$ &  $0.02$ &  $0.69$ &  $0.07$ &  $\textbf{ 0.81 }$ &  $0.01$ &  $0.82$ &  $0.01$ &  $0.80$ &  $0.01$ &  $0.95$ &  $0.04$ &  $0.97$ &  $0.02$ &  $0.98$ &  $0.03$ &  $0.98$ \\
         VQ-GAN f16-kl &  $0.34$ &  $0.28$ &  $0.34$ &  $0.04$ &  $23.51$ &  $0.07$ &  $0.08$ &  $0.09$ &  $0.11$ &  $0.13$ &  $0.16$ &  $0.02$ &  $1.00$ &  $0.02$ &  $0.68$ &  $0.07$ &  $0.80$ &  $0.01$ &  $0.80$ &  $0.01$ &  $0.79$ &  $0.01$ &  $0.96$ &  $0.04$ &  $0.98$ &  $0.02$ &  $0.98$ &  $0.03$ &  $0.98$ \\
         VQ-GAN f32-kl &  $0.27$ &  $0.22$ &  $0.28$ &  $0.05$ &  $21.80$ &  $0.09$ &  $0.09$ &  $0.10$ &  $0.12$ &  $0.14$ &  $0.17$ &  $0.02$ &  $1.00$ &  $0.02$ &  $0.63$ &  $0.08$ &  $0.77$ &  $0.01$ &  $0.77$ &  $0.01$ &  $0.76$ &  $0.01$ &  $0.95$ &  $0.04$ &  $0.97$ &  $0.03$ &  $0.97$ &  $0.03$ &  $0.97$ \\
         \bottomrule
    \end{tabular}
    \caption{Preliminary evaluation of the ImageNet pre-trained codebooks from Figure 3 in the main paper represented in numerical form. The different rows show different pretrained Codebooks, columns indicate different datasets where it is indicated in brackets whether the images of the datasets are autoencoded or the segmentation maps are autoencoded to determine the upper performance bounds with respective codebooks.}
    \label{tab:codebook_eval2}
\end{table*}
}

\begin{table}[]
    \centering
    \begin{tabular}{lcc}
        \toprule
        Hyperparameter & VQ-GAN & Transformer \\
        \midrule
        learning rate & $4.5e-06$ & $4.5e-06$ \\
        batch size & $96$ & $4$ \\
        codebook size & $16384$ & $16384 + 2$ \\
        codebook enc. resolutions & 1, 1, 2, 2, 4 & N/A \\
        $^\llcorner$residual blocks per level & 2 & N/A \\
        block size & N/A & $4,500$ \\
        \# encoder layers & N/A & $6$ \\
        \# attention heads & N/A & $16$ \\
        embdding dimension & N/A & $1,408$ \\
        dropout ratio (embedding) & N/A & $0.1$ \\
        dropout ratio (residual) & N/A & $0.1$ \\
        dropout ratio (attention) & N/A & $0.1$ \\
        hardware & $4\times$NVIDIA & $4\times$NVIDIA \\
        &  A100 40GB &  A100 40GB \\
        \bottomrule
    \end{tabular}
    \caption{Overview of the hyperparamters for both the VQ-GAN models and the GPT2 transformers we train.}
    \label{tab:hyperparams}
\end{table}
\subsection{Qualitative compositional predictions}
\begin{figure*}
    \centering
    \begin{tabular}{l}
    Context compositional task sequence \\    \includegraphics[width=\textwidth,page=4]{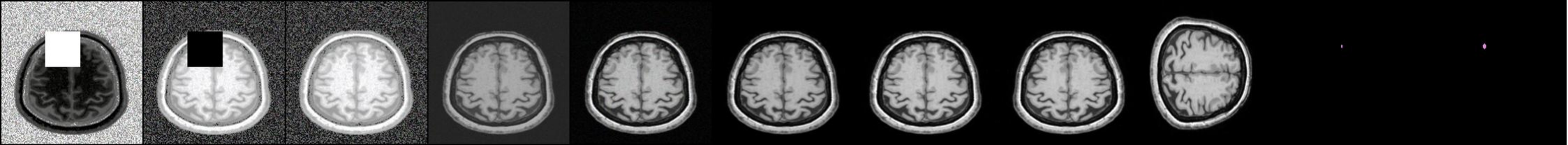} \\
    Query$\phantom{0000}$Predicted sequential output \\    \includegraphics[width=\textwidth,page=5]{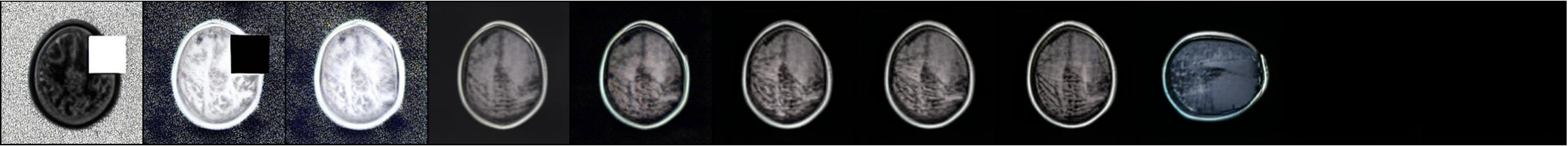} \\
    Query$\phantom{0000}$Ground-truth compositional task sequence \\
    \includegraphics[width=\textwidth,page=6]{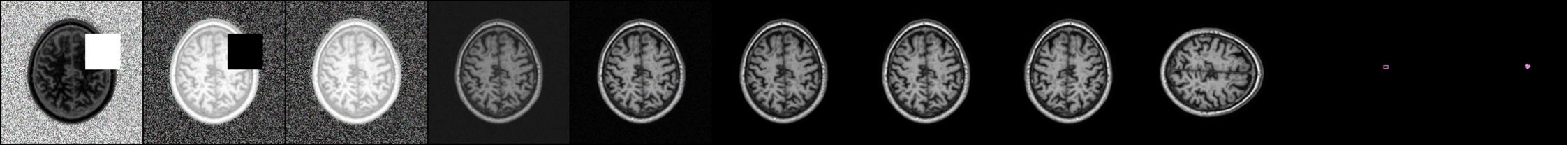} \\
    \midrule \\
    Context compositional task sequence \\    \includegraphics[width=\textwidth,page=8]{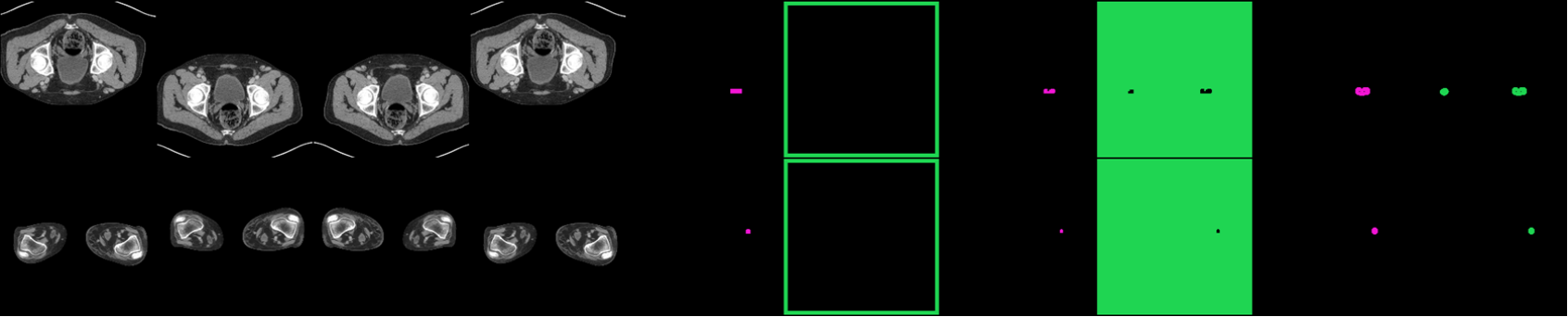} \\
    Query$\phantom{00000}$Predicted sequential output \\    \includegraphics[width=\textwidth,page=9]{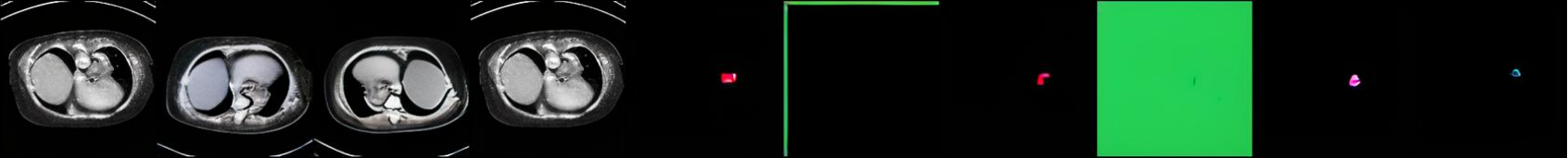} \\
    Query$\phantom{00000}$Ground-truth compositional task sequence \\
    \includegraphics[width=\textwidth,page=10]{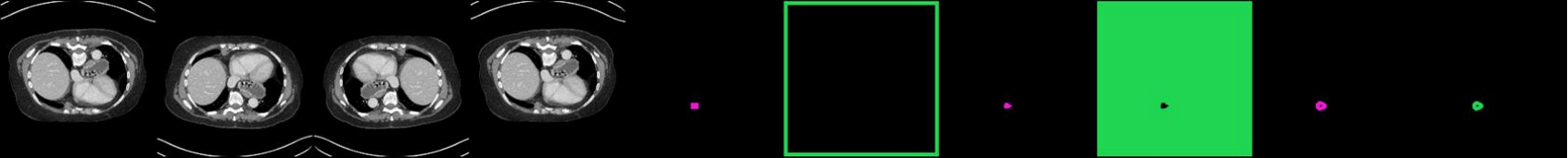} \\
    \end{tabular}
    \caption{Additional compositional contexts, predicted task sequences and the associated ground-truth task sequences.}
    \label{fig:additional_qualitative_results}
\end{figure*}

\begin{figure*}
    \centering
    \begin{tabular}{l}
    Context compositional task sequence \\    \includegraphics[width=\textwidth,page=12]{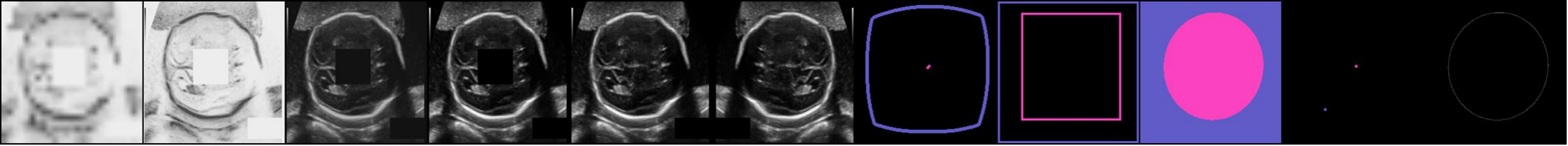} \\
    Query$\phantom{0000}$Predicted sequential output \\    \includegraphics[width=\textwidth,page=13]{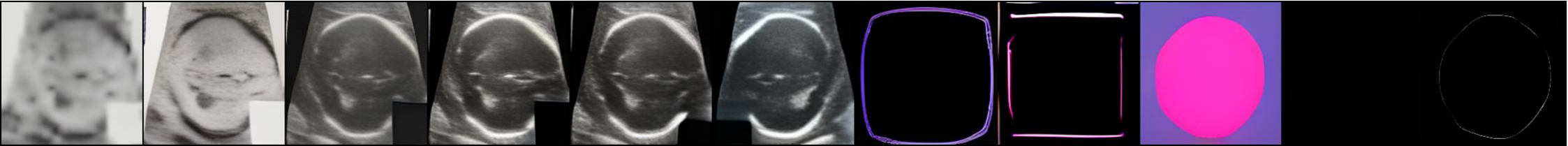} \\
    Query$\phantom{0000}$Ground-truth compositional task sequence \\
    \includegraphics[width=\textwidth,page=14]{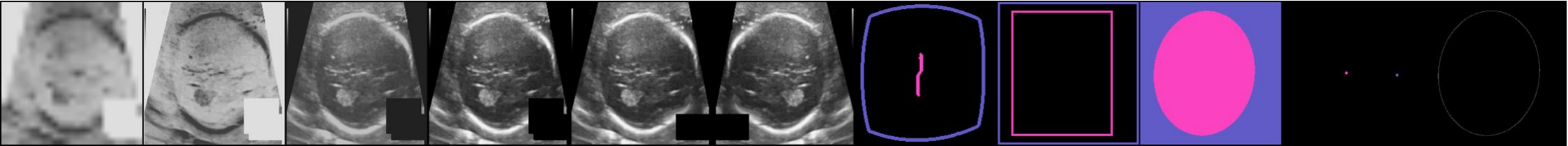} \\
    \midrule \\
    Context compositional task sequence \\    \includegraphics[width=\textwidth,page=16]{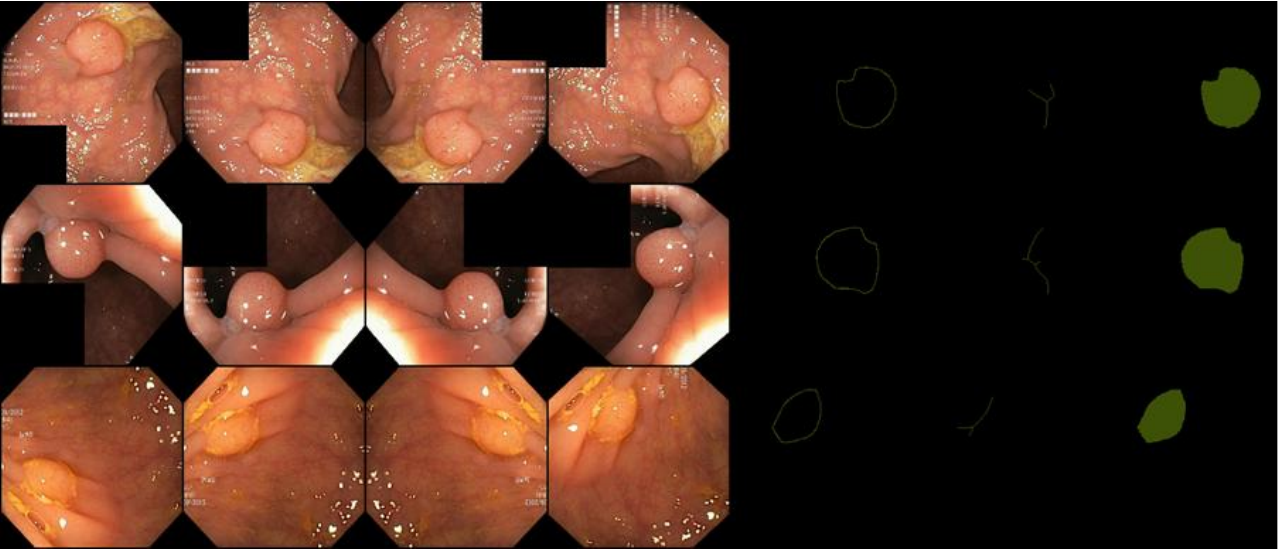} \\
    Query$\phantom{0000000000}$Predicted sequential output \\    \includegraphics[width=\textwidth,page=17]{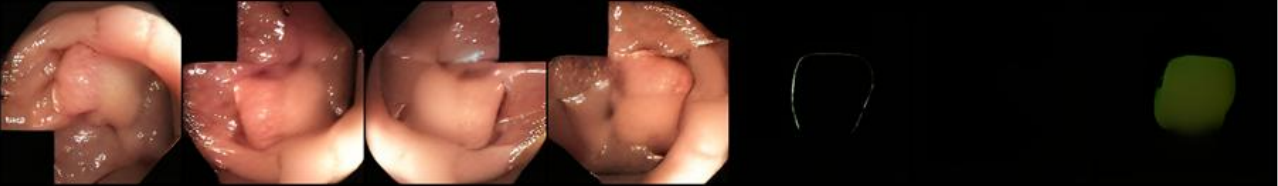} \\
    Query$\phantom{0000000000}$Ground-truth compositional task sequence \\
    \includegraphics[width=\textwidth,page=18]{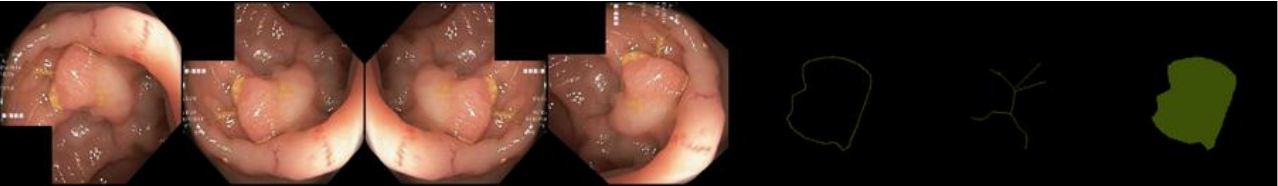} \\
    \end{tabular}
    \caption{Additional compositional contexts, predicted task sequences and the associated ground-truth task sequences.}
    \label{fig:additional_qualitative_results2}
\end{figure*}

In~\Cref{fig:additional_qualitative_results} and~\Cref{fig:additional_qualitative_results2} we display additional compositional task sequences and associated predictions by the visual in-context transformer.
We can observe, that the visual in-context learner, even though the compositional tasks are quite complex and encompass long sequences, can coherently follow the instructions in the context task sequence.
Of course, we can also observe inaccuracies, such that certain structures can not be predicted coherently, such as detailed structures in the brain in~\Cref{fig:additional_qualitative_results} (top).

Interestingly, when inspecting the image modalities of both~\Cref{fig:additional_qualitative_results} and~\Cref{fig:additional_qualitative_results2}, the wide range of modalities, \ie, magnetic resonace imaging, computed tomography, ultrasound, and even endoscopy images can be captured nicely.

In some cases, we see, that the visual in-context learner can capture the distribution of the expected output, \eg, the segmentation in the last two columns of~\Cref{fig:additional_qualitative_results}, in the example at the bottom, but the predicted structures are located at the wrong position.
This might be a hint that in case the model is not able to connect the image pattern to the semantic structure, as defined in the context set, it hallucinates a wrong but overall plausible-looking output.
Fine, small structures in medical imaging datasets might elevate this problem which also highlights the challenging setting.

\section{Limitations}
Despite the promising results, the approach has several limitations that need to be addressed in future work.
The granularity of what the codebooks can represent limits their ability to recover fine-grained structures and, while color remapping augmentation improves generalization, the model remains sensitive to color variations where classes are encoded in very similar colors (\eg, different shades of blue encoding different classes).

Ensuring consistency across intermediate outputs of the visual in-context learner is a remaining challenge as well.
The model currently may produce misaligned sub-task predictions, affecting the overall coherence in the output sequence.
This also ties into the qualitative observation that with longer sequences the intermediate outputs degrade successively (\eg,~\Cref{fig:additional_qualitative_results}, second row where late in the sequence, the predicted brain scan is tinted blue).

The synthetic task generation pipeline, although effective, may not fully capture the complexity and diversity of real-world tasks which might impact the model's generalization capabilities.
Exploring the effects of adapting this task generation pipeline from a data-centric perspective would be beneficial.

Balancing training data between imaging data and task outputs is crucial but challenging, especially with diverse datasets with different amounts of samples and the different generative-, transformation- and discriminative tasks.

Evaluation metrics such as IoU, F-1 Score, MAE, RMSE, and PSNR, while useful, may not fully capture the nuances of complex task sequences, necessitating more comprehensive evaluation frameworks.
Here, efforts towards designing a compositional medical task benchmark are needed which encompasses a wide variety of compositional task prompts and associated evaluation schemes.

Finally, while the model shows promising results on complex, compositional task sequences, there is still much room for improvement when moving to out-of-domain sequences, \ie, when moving beyond training datasets.
One elevating factor for the presented approach could lie in model- and data scale, as visual in-context learners that were trained on larger datasets, with higher parameter counts~\cite{bai2023sequential} exhibited strong out-of-domain capabilities.

Addressing these limitations will be crucial for advancing visual in-context learning and enabling more robust, adaptable models for a wide range of applications. Future work should focus on enhancing codebooks, improving training strategies, expanding the synthetic task pipeline, and developing comprehensive evaluation frameworks.

\section{Utilized open source code and libraries}
We heavily build on the two codebases~\footnote{\url{https://github.com/CompVis/taming-transformers}}~\footnote{\url{https://github.com/CompVis/latent-diffusion}} and their opensource models.
The GPT-2 implementation is based on the repository of minGPT~\footnote{\url{https://github.com/karpathy/minGPT/}}.
Further, we make heavy use of libraries such as Pytorch~\cite{NEURIPS2019_9015}, Pandas~\cite{mckinney2010data} and sklearn~\cite{scikit-learn}.

\section{Impact statement}
This paper presents work whose goal is to explore machine learning techniques for solving compositional visual tasks from the field of medical image analysis.
Specifically, our goal is to enable users to specify vision-task pipelines without the requirement of programming knowledge or model re-training.
This goal inherently means, that a broader audience could be able to use the developed models, which may include users with malicious intent.
As the current models are not able to produce results that are on a level to be used in medical practice and mainly serve to advance research, the factor of misuse may be of relatively low concern.

We utilize a broad set of medical datasets for this exploration and train models on it.
As such, these models may reflect the biases within these datasets, such as gender-, age- or ethnicity imbalance.